\documentclass{article}
\usepackage{colm2024_conference}
\usepackage{scrextend}
\usepackage{booktabs}
\usepackage{graphicx}
\usepackage{enumitem}
\usepackage{algorithm}
\usepackage{algpseudocode}
\usepackage{natbib}
\usepackage{makecell}
\usepackage{array}
\usepackage{amsmath}
\usepackage{amssymb}
\usepackage{amsfonts}
\usepackage{multirow}
\usepackage{caption}
\usepackage{subcaption}
\usepackage{tabularx}
\usepackage{colortbl}
\usepackage{float}
\usepackage{placeins}
\usepackage{hyperref}
\usepackage[utf8]{inputenc}
\usepackage[T1]{fontenc}
\usepackage{xspace}
\usepackage{url}
\usepackage{nicefrac}
\usepackage{textcomp}
\usepackage{siunitx}
\usepackage{listings}
\usepackage{adjustbox}

\renewcommand{\arraystretch}{1.18}
\algrenewcommand\alglinenumber[1]{\scriptsize #1:}

\newcommand{\model}{S1-Omni-Image\xspace}
\newcommand{\sonevl}{S1-VL-32B\xspace}
\newcommand{\qwenedit}{Qwen-Image-Edit-2511\xspace}

\title{S1-Omni-Image: A Unified Model for Scientific Image \\
Understanding, Generation, and Editing}

\author{
\bf XScience Lab\\
Wenge AI}

\begin{document}

\maketitle

\begin{abstract}
We present \model, an open-weight unified multimodal model for scientific image understanding, generation, and editing. Unlike general-purpose image generation models, scientific image tasks require not only high-fidelity synthesis, but also robust understanding of scientific semantics, structural relations, domain knowledge, and task intent. To this end, \model builds on the scientific multimodal reasoning backbone \sonevl and couples its understanding capability with an image generation module under a unified \textit{think-before-generate} paradigm. Given a user instruction, the model first produces a task-oriented reasoning trace, a textual answer, and a task special token; their hidden states are then injected into the generation module to condition image generation or editing. \model supports scientific image understanding, generation, and editing in a unified framework. For generation, it focuses on scientific illustrations and text rendering, including logical diagrams, relational comparisons, data charts, and realistic scientific visualizations. For editing, it casts segmentation and other domain-specific vision tasks as native image editing problems, enabling multi-turn illustration editing, medical and geographic image segmentation, medical image translation, and scientific image super-resolution. We construct SciGenEdit, a 314K-sample training dataset, and release the model weights, inference code, and SciGenEdit-10K. Experiments show that \model substantially improves scientific image generation and editing while preserving the scientific image understanding capability inherited from \sonevl. It outperforms open-source models on GenExam and TechImage-Bench, achieves state-of-the-art results on four editing benchmarks including MSD, cigRockSEM, SynthRAD2025, and IXI, and maintains stable performance on scientific image understanding evaluations.
\end{abstract}

\begin{figure}[h]
    \centering
    \includegraphics[width=0.96\linewidth]{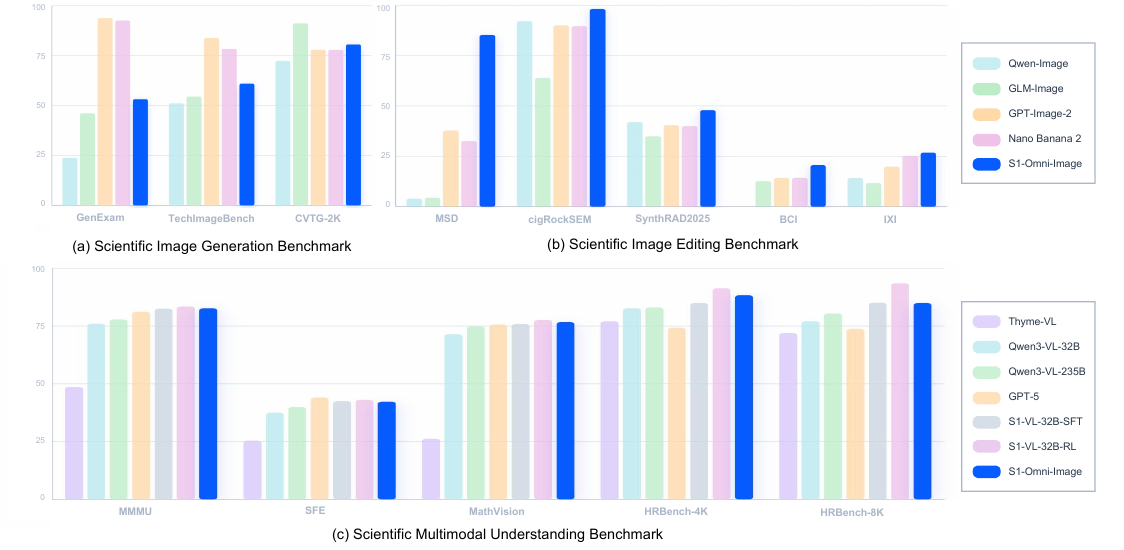}
    \caption{Overall performance of \model on scientific image generation (a), scientific image editing (b), and scientific image understanding (c). For visualization, the MSD and cigRockSEM scores in the image editing panel are multiplied by 100; bar heights should not be interpreted as directly comparable across different benchmarks.}
    \label{fig:overview}
\end{figure}

\begin{figure}[t!]
    \centering
    \vspace*{-0.7cm}
    \includegraphics[width=1.0\textwidth]{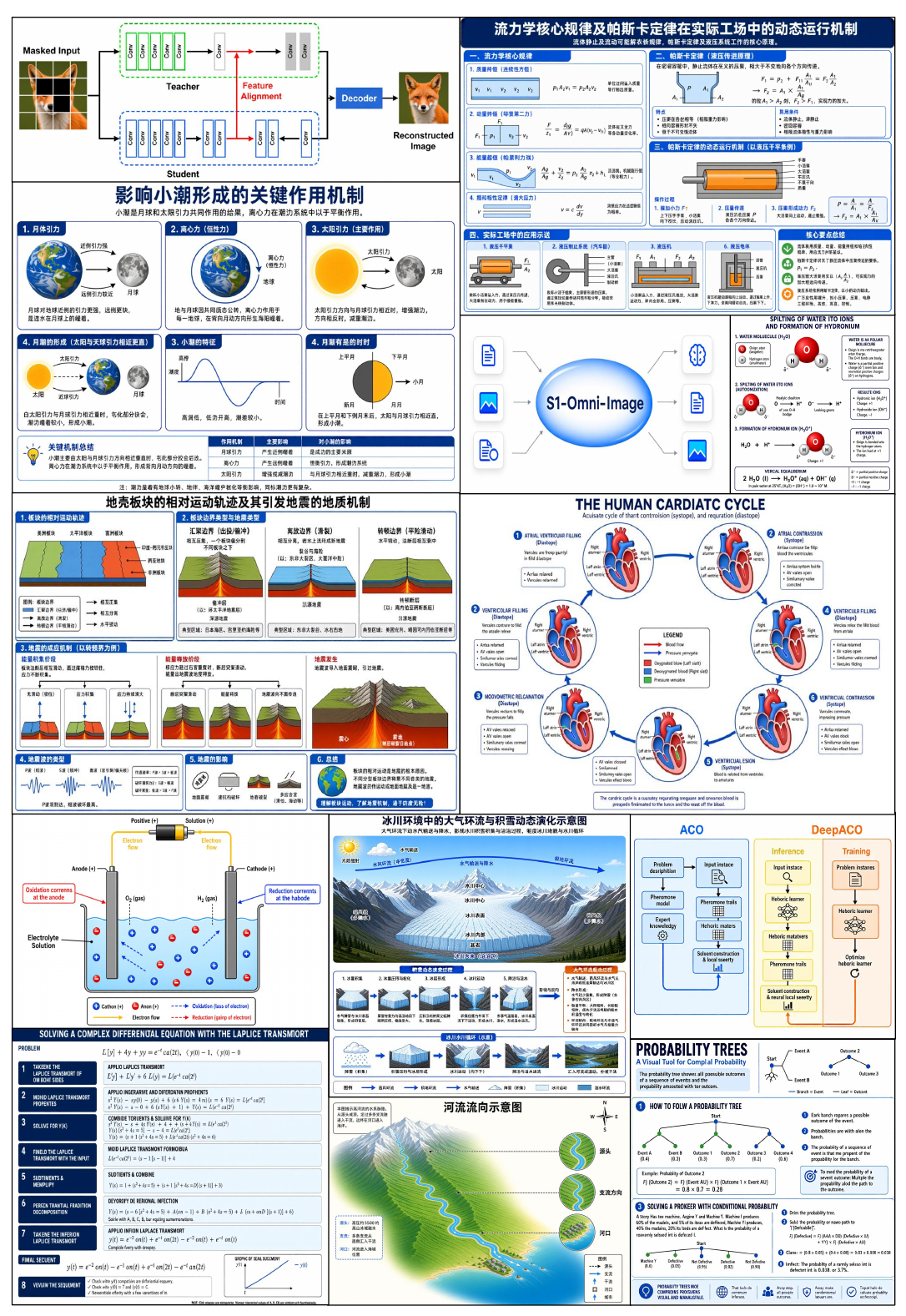}
    \caption{Capability overview of \model on scientific image generation. 
    The examples demonstrate the model's ability to generate scientific illustrations from textual instructions, including structured framework diagrams, mechanism illustrations, relational comparisons, data charts, and realistic scientific visualizations.}
    \label{fig:intro_cases_gen}
\end{figure}

\clearpage

\begin{figure}[t!]
    \centering
    \vspace*{-0.7cm}
    \includegraphics[width=\textwidth]{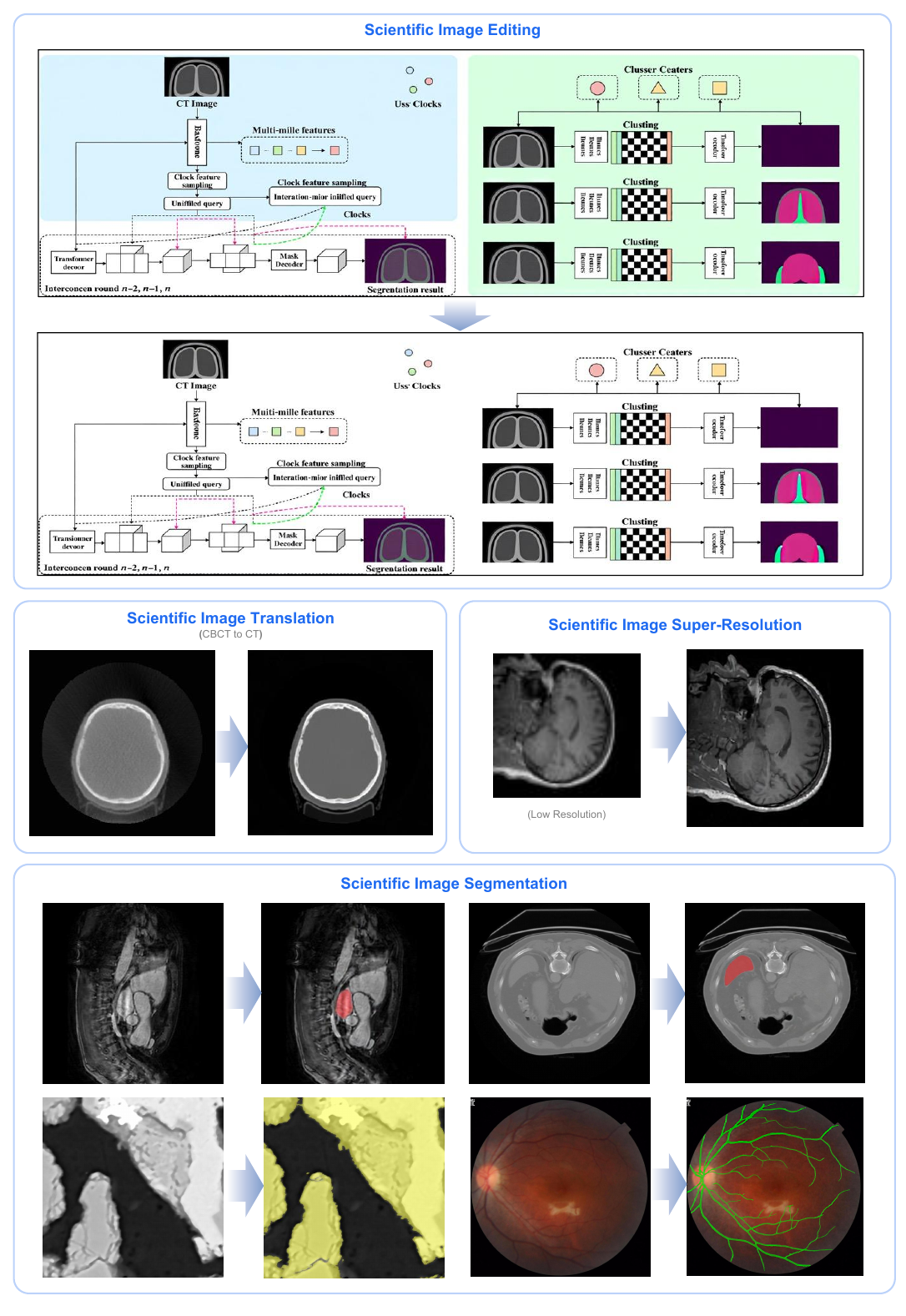}
    \caption{Capability overview of \model on scientific image editing. 
    The examples cover scientific image segmentation, scientific illustration editing, scientific image super-resolution, and scientific image translation such as CBCT-to-CT conversion. These tasks can be handled as instruction-conditioned image editing.}
    \label{fig:intro_cases_edit}
\end{figure}

\clearpage

\section{Introduction}
\label{sec:intro}

Scientific research relies heavily on visual communication. Flowcharts, framework diagrams, mechanism illustrations, and conceptual figures organize complex knowledge, describe experimental procedures, and communicate model designs. Medical images, microscopy images, material images, and geological images further serve as direct carriers of structural recognition, regional analysis, and cross-modal transformation. Scientific illustrations have been recognized as an important complement to scientific communication, helping readers grasp contributions more efficiently and reducing the risk of misinterpretation~\citep{fytas2021makesscientificpaperaccepted,kim2022seeing}. With the rapid development of multimodal models and image generation models, automated scientific image generation and editing are becoming increasingly important research directions~\citep{Zhu2026AutoFigure,Rodriguez2023FigGen,Wu2025QwenImage}.

Scientific images differ fundamentally from ordinary natural images or artistic images. Their central requirements are not limited to visual realism or aesthetic quality; they also include semantic correctness, structural clarity, relational consistency, and controllability under domain-specific constraints. For example, a scientific framework diagram must correctly express modules, arrows, and hierarchical relations; a medical segmentation output must accurately localize anatomical structures or lesions; medical image translation and super-resolution must improve image quality or convert modality while preserving critical structural information. Recent studies on scientific illustration generation and evaluation further show that scientific image generation must simultaneously consider content faithfulness, structural fidelity, readability, and visual design quality, while long scientific context, complex topology, and text rendering substantially increase the difficulty~\citep{Zhu2026AutoFigure,chang2025sridbenchbenchmarkscientificresearch,liu2025improving,du2025textcrafter}. Scientific image generation and editing are therefore best understood as reasoning-driven visual generation problems.

Existing approaches remain insufficient for this setting. On the one hand, multimodal large language models have made substantial progress in image understanding, document understanding, and scientific reasoning~\citep{qwenvl,Chen2023InternVL,bai2025qwen3,bai2025intern,singh2025openai,comanici2025gemini}, yet most of them primarily produce textual answers and cannot directly perform high-quality image generation or editing. On the other hand, open-source image generation and editing models achieve strong results in general domains~\citep{Ho2020DDPM,Rombach2022HighResolution,Wu2025QwenImage,liu2025step1x,ye2025imgedit}, but their training objectives mainly target natural images, creative design, and general editing. When applied to scientific images, they often suffer from structural errors, unstable text rendering, inconsistent scientific semantics, and drift across multi-turn editing. Moreover, scientific illustration generation systems often rely on agents or tool-chain pipelines~\citep{Zhu2026AutoFigure,belouadi2023automatikz,belouadi2024detikzify,belouadi2025tikzero}, while medical segmentation, medical image translation, and super-resolution are usually handled by separate task-specific models~\citep{Isensee2021nnUNet,Han2025AllInOneMedicalI2I,Saharia2021SR3}. A unified open-weight model for scientific image understanding, generation, and editing is still missing.

Unified multimodal models provide a promising technical path by integrating understanding, generation, and editing within a single framework~\citep{Sun2023Emu,Sun2023Emu2,wang2024emu3,Xie2024ShowO,Ge2024SEEDX,Wu2024Janus,Chen2025JanusPro,deng2025bagel,liu2025tuna,lance2026}. However, most unified models are still primarily designed for general image domains and lack systematic optimization for scientific semantics, professional structures, region-level editing, and cross-task scientific image transformations. More importantly, complex scientific images often require the model to understand the task intent, infer key structures, plan the visual expression, and then execute generation or editing. Directly mapping a user prompt to an image makes it difficult to robustly handle long context, domain concepts, and multi-step editing.

We therefore propose \model, a unified multimodal model for scientific image understanding, generation, and editing. Instead of directly translating prompts into images, \model conditions image generation and editing on reasoning representations produced by a scientific image understanding model. Specifically, \model is initialized from \sonevl~\citep{li2026s1}, which serves as the scientific multimodal reasoning backbone. It understands and reasons over text, images, and image-text combinations, and injects the resulting hidden states into an MMDiT image generation and editing module through a reasoning-to-diffusion alignment layer. This design instantiates a \textit{think-before-generate} paradigm in the context of scientific image generation and editing. We release the model weights, inference service code, and SciGenEdit-10K under the Apache 2.0 license to facilitate reproduction and future research. Figure~\ref{fig:overview} summarizes the overall performance of \model across scientific image generation, scientific image editing, and scientific image understanding benchmarks, while Figures~\ref{fig:intro_cases_gen} and~\ref{fig:intro_cases_edit} illustrate representative capabilities of \model on scientific image generation and scientific image editing, respectively.

Our main contributions are as follows:
\begin{itemize}[leftmargin=*]
    \item We introduce \model, an open-weight unified multimodal model for scientific image understanding, generation, and editing. It formulates scientific image generation, image editing, segmentation, image translation, and super-resolution within a unified framework.
    \item We design a unified \textit{think-before-generate} paradigm for diverse scientific image tasks. The model first performs task-oriented scientific reasoning and then uses route tokens to trigger generation or editing. A reasoning-to-diffusion alignment module maps reasoning hidden states into the MMDiT conditioning space, enabling reasoning-guided control over scientific semantics and visual details.
    \item We construct the SciGenEdit dataset and release model weights, inference service code, and the SciGenEdit-10K subset. We conduct systematic evaluation on scientific image generation, scientific image editing, and scientific image understanding benchmarks.
\end{itemize}

\section{Related Work}
\label{sec:related}

\subsection{Image Understanding Models}
\label{sec:related_understanding}

Multimodal large language models typically combine a visual encoder, a cross-modal connector, and a language-model backbone to process image-text inputs. Representative systems include LLaVA, Qwen-VL, InternVL, and Gemini~\citep{Liu2023VisualInstruction,Liu2023ImprovedBaselines,qwenvl,Chen2023InternVL,team2023gemini,comanici2025gemini}. These models have achieved strong performance on visual question answering, document understanding, chart analysis, OCR, and multimodal reasoning, and have gradually expanded from single-image question answering to long-context, multi-image, and complex reasoning scenarios. Nevertheless, their outputs remain primarily textual: visual information is usually encoded at the input side, while the model itself does not directly perform high-fidelity image generation or instruction-based image editing.

Scientific domains impose stronger requirements on multimodal understanding. Images often contain dense text, formulas, charts, microscopic structures, medical scans, or remote-sensing textures, and reasoning frequently requires domain knowledge together with local visual evidence. Intern-S1 and GMAI-MMBench advance scientific and medical multimodal modeling and evaluation~\citep{bai2025intern,chen2024gmai}, while Qwen3-VL further improves general multimodal reasoning capability~\citep{bai2025qwen3}. S1-VL introduces Thinking-with-Images for scientific reasoning, enabling the model to actively crop, zoom, or annotate images during reasoning and thereby improving high-resolution, fine-grained scientific image understanding~\citep{li2026s1,zhang2025thyme,su2025thinking}. \model inherits this scientific understanding capability and further uses it to drive image generation and editing.

\subsection{Image Generation Models}
\label{sec:related_generation}

Image generation has evolved from GANs and VQ/autoregressive models to diffusion models and flow-based models. Latent diffusion and DiT/MMDiT architectures have substantially improved high-resolution generation quality and text-conditioned controllability~\citep{Ho2020DDPM,Rombach2022HighResolution,Wu2025QwenImage}. Recent models such as Qwen-Image further emphasize complex prompt following, multilingual text rendering, long-text layout, and integrated generation-editing ability~\citep{Wu2025QwenImage,du2025textcrafter}. These models perform well on natural images, creative content, design, and general visual generation, but their data and evaluation objectives are not specifically tailored to scientific images.

Scientific image generation places greater emphasis on structured expression, symbolic relations, textual annotation, layout logic, and domain-semantic consistency. Existing work explores this direction from multiple angles. FigGen studies scientific figure generation from short text or captions~\citep{Rodriguez2023FigGen}; AutoFigure/FigureBench targets long-context scientific illustration design and emphasizes extracting methodological structure from paper context to produce publication-ready figures~\citep{Zhu2026AutoFigure}; SridBench focuses on structural and semantic evaluation of scientific illustration generation~\citep{chang2025sridbenchbenchmarkscientificresearch}. In addition, AutoTikZ, DeTikZify, and TikZero use TikZ or executable graphics programs as intermediate representations to improve control over geometry and symbolic layout~\citep{belouadi2023automatikz,belouadi2024detikzify,belouadi2025tikzero}. These approaches demonstrate the importance of explicit planning and structural constraints, but most rely on agents, code generation, or post-processing, rather than an end-to-end open unified model.

\subsection{Image Editing Models}
\label{sec:related_editing}

General image editing models have been developed around instruction following, local repainting, attention control, conditional control, and editing-data construction. Prompt-to-Prompt, DiffEdit, ControlNet, and InstructPix2Pix advance controllable editing from the perspectives of cross-attention manipulation, mask-guided editing, external condition injection, and natural-language instruction editing~\citep{Hertz2022PromptToPrompt,Couairon2022DiffEdit,Zhang2023ControlNet,Brooks2022InstructPix2Pix}. More recent work such as Step1X-Edit and ImgEdit constructs large-scale editing datasets and benchmarks, improving open-domain instruction editing, detail preservation, and multi-category editing~\citep{liu2025step1x,ye2025imgedit}. However, general editing models typically assume natural or design images, and their evaluation mainly focuses on visual quality, local consistency, and instruction following.

Scientific image editing is subject to more complex constraints. It includes editing modules, arrows, labels, and layouts in scientific illustrations, as well as domain tasks such as medical image segmentation, scientific image segmentation, medical image translation, and image super-resolution. Traditionally, these tasks are handled by task-specific models: medical image segmentation relies heavily on U-Net, nnU-Net, and their variants~\citep{Ronneberger2015UNet,Isensee2021nnUNet}; medical image translation often uses GAN-based or diffusion-based translation methods~\citep{Isola2017Pix2Pix,Han2025AllInOneMedicalI2I}; super-resolution relies on specialized reconstruction networks or diffusion models~\citep{Saharia2021SR3}. Although effective on specific datasets, these methods do not share a unified input-output interface and are difficult to extend to open instructions, multi-turn interaction, and cross-domain generalization. Scientific image editing therefore calls for a modeling paradigm that unifies segmentation, translation, super-resolution, and scientific illustration editing.

\subsection{Unified Multimodal Models}
\label{sec:related_unified}
Unified multimodal models aim to process and generate multiple modalities within a single framework, spanning text, images, video, audio, and interleaved multimodal content. When such models emphasize broad task coverage across perception, reasoning, and generation, they are also often described as omni models. One line of work adopts unified token modeling; for example, Emu3 explores next-token prediction as a single objective for image, text, and video tokens~\citep{wang2024emu3}. Another line combines autoregressive language modeling with diffusion or flow matching to support image understanding, image generation, and instruction-based editing, as in Show-o, SEED-X, Janus, BAGEL, LANCE, and TUNA~\citep{Xie2024ShowO,Ge2024SEEDX,Wu2024Janus,Chen2025JanusPro,deng2025bagel,liu2025tuna,lance2026}.

Despite their broader capability boundaries, most existing unified models are still optimized mainly for general images, design, and open-domain creation. They lack targeted training for scientific image structures, scientific semantics, region-level editing, and cross-task scientific image transformation. Scientific image tasks require the model not only to produce visually plausible outputs, but also to infer task intent, reason over key structures, plan visual expression, and preserve domain semantics during editing. Unlike prior unified or omni models, \model uses a scientific multimodal reasoning backbone as the conditional source and injects explicit reasoning representations into generation and editing through reasoning-to-diffusion alignment. The model first produces task-relevant reasoning and then uses the scientific reasoning representation to drive visual synthesis.

\section{Model Architecture}
\label{sec:model}

\subsection{Overview}
\label{sec:model_overview}

\begin{figure}[t!]
    \centering
    \includegraphics[width=0.96\linewidth]{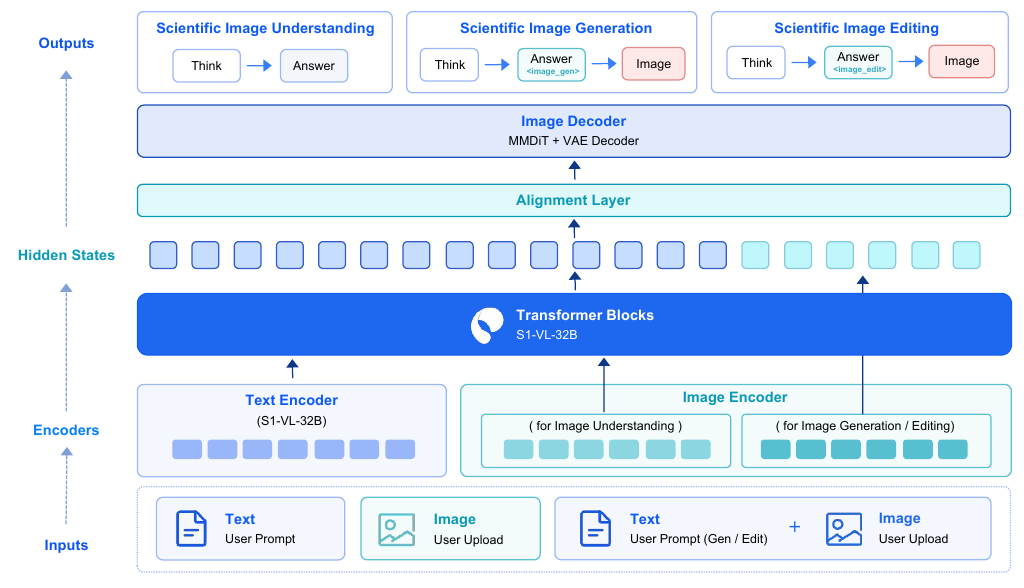}
    \caption{Architecture of \model. The model uses \sonevl as the scientific multimodal reasoning backbone and injects reasoning representations into an MMDiT image generation and editing module through a reasoning-to-diffusion alignment layer. Understanding outputs are generated by the VLM text branch, while the image decoder is used only for generation and editing.}
    \label{fig:model}
\end{figure}

The overall architecture of \model is shown in Figure~\ref{fig:model}. The model supports text, image, and image-text inputs, and produces textual answers, reasoning traces, generated images, or edited images depending on the task. Given a text instruction $x$ and an optional input image $I_{\mathrm{in}}$, \model first uses the Stage-I-tuned \sonevl to autoregressively generate an explicit reasoning sequence $r$, a textual answer $a$, and a task special token $s$. Here, $s=\texttt{<image\_gen>}$ denotes image generation and $s=\texttt{<image\_edit>}$ denotes image editing. During the same generation process, the model retains the hidden states corresponding to the user input, reasoning trace, textual answer, and task special token. These representations are aligned to the conditioning space of the diffusion generation module and used to guide image generation or editing.

Algorithm~\ref{alg:tbg} summarizes the inference procedure. Here, $r$ is the explicit reasoning trace, $a$ is the textual answer, $s$ is the task special token, $\tilde{x}$ is the full context composed of the input and generated response, $H$ denotes the hidden states collected during the same generation process, $C$ denotes the aligned diffusion-conditioning representation, $z_t$ is the noisy latent at diffusion time step $t$, and $\hat{I}$ is the generated or edited image. For text-to-image generation, $I_{\mathrm{in}}$ is empty; for image editing, $I_{\mathrm{in}}$ is the input image to be edited.

The special token also serves as a task router:
\begin{itemize}[leftmargin=*]
    \item \texttt{<image\_gen>}: route to the image generation tasks;
    \item \texttt{<image\_edit>}: route to the image editing tasks.
\end{itemize}
During training and inference, image generation samples follow the format ``user text instruction $\rightarrow$ assistant outputs \texttt{<think>}, explicit reasoning, \texttt{</think>}, textual answer, and \texttt{<image\_gen>}''. Image editing samples additionally include an input image on the user side, following ``input image + user editing instruction $\rightarrow$ assistant outputs \texttt{<think>}, explicit reasoning, \texttt{</think>}, textual answer, and \texttt{<image\_edit>}''. Generation and editing therefore share the same reasoning-and-answer prefix and are distinguished only by the existence of an input image and the final special token.

\subsection{Scientific Multimodal Reasoning Backbone}
\label{sec:model_backbone}

\model uses a model initialized from \sonevl and further tuned in Stage I as the scientific multimodal reasoning backbone. Given input $x$ and optional image $I_{\mathrm{in}}$, \sonevl autoregressively generates a response $y=[r,a,s]$ consisting of an explicit reasoning sequence, a textual answer, and a task special token. During this generation process, we collect multimodal contextual representations for input tokens, visual tokens, the explicit reasoning sequence, and the task special token:
\begin{equation}
H = \{h_1, h_2, \ldots, h_n\}
= \mathrm{HiddenStates}_{\mathrm{VL}}(x,I_{\mathrm{in}},y),
\end{equation}
where $h_i \in \mathbb{R}^{d_{\mathrm{LLM}}}$ and $d_{\mathrm{LLM}}=5120$. These hidden states jointly encode input text semantics, image content, task intent, and reasoning-process information. The response itself is generated with the standard autoregressive language modeling objective:
\begin{equation}
p(y \mid x, I_{\mathrm{in}}) = \prod_{t=1}^{T} p(y_t \mid y_{<t}, x, I_{\mathrm{in}}),
\end{equation}
where $y$ denotes the response composed of the reasoning sequence, textual answer, and task special token.

\begin{algorithm}[t!]
\caption{Think-Before-Generate Inference of \model}
\label{alg:tbg}
\scriptsize
\begin{algorithmic}[1]
\State Given a user instruction $x$ and an optional input image $I_{\mathrm{in}}$.
\State Use the Stage-I-tuned \sonevl to autoregressively generate reasoning $r$, answer $a$, and task special token $s$, while collecting hidden states from this generation pass:
\Statex \hspace{\algorithmicindent}$(r,a,s), H \leftarrow \mathrm{Decode}_{\mathrm{VL}}(x, I_{\mathrm{in}})$.
\State Define the full context containing the reasoning result:
\Statex \hspace{\algorithmicindent}$\tilde{x}=[x,r,a,s]$.
\State Project hidden states into the MMDiT conditioning space:
\Statex \hspace{\algorithmicindent}$C=f_{\mathrm{align}}(H)$.
\State Route to image generation or image editing according to $s$, and obtain the final image through the diffusion module:
\Statex \hspace{\algorithmicindent}$\hat{I}=f_{\mathrm{diff}}(z_t,t,C,I_{\mathrm{in}})$.
\end{algorithmic}
\end{algorithm}

\subsection{MMDiT-based Image Generation and Editing Module}
\label{sec:model_mmdit}

The image generation and editing module of \model is based on the MMDiT architecture. It reuses the VAE encoder, VAE decoder, and MMDiT weights of the public Qwen-Image model; the specific version used in this work is \qwenedit. The original condition encoder of Qwen-Image is Qwen2.5-VL, and its MMDiT expects conditioning representations with dimensionality 3584. Different from diffusion models based on noise prediction, \model follows the flow-matching paradigm of Qwen-Image, where MMDiT directly predicts the velocity field from a noise latent to a target image latent. Given a target image $I$, the VAE encoder maps it to the latent space:
\begin{equation}
z_{\mathrm{data}} = E_{\mathrm{VAE}}(I).
\end{equation}
We then sample a noise latent from a standard Gaussian distribution:
\begin{equation}
z_{\mathrm{noise}} \sim \mathcal{N}(0,\mathbf{I}).
\end{equation}
At time step $t \in [0,1]$, flow matching constructs an intermediate latent by linear interpolation:
\begin{equation}
z_t = t z_{\mathrm{data}} + (1-t) z_{\mathrm{noise}},
\end{equation}
whose target velocity field is
\begin{equation}
v_t = \frac{d z_t}{dt} = z_{\mathrm{data}} - z_{\mathrm{noise}}.
\end{equation}
The MMDiT module takes time step $t$, intermediate latent $z_t$, and the aligned conditioning representation $C$ as input and predicts the flow-matching velocity:
\begin{equation}
\hat{v} = v_{\theta}(z_t,t,C).
\end{equation}
The training objective minimizes the mean squared error between the predicted velocity and target velocity:
\begin{equation}
\mathcal{L}_{\mathrm{FM}} =
\mathbb{E}_{z_{\mathrm{data}}, z_{\mathrm{noise}}, t}
\left[
\left\|
v_{\theta}(z_t,t,C) -
\left(z_{\mathrm{data}} - z_{\mathrm{noise}}\right)
\right\|_2^2
\right].
\end{equation}

\subsection{Reasoning-to-Diffusion Alignment Layer}
\label{sec:model_alignment}

The reasoning-to-diffusion alignment layer is the key component that connects \sonevl with the MMDiT generation module. Since the hidden states of \sonevl lie in the LLM representation space, whereas Qwen-Image MMDiT expects inputs in the Qwen2.5-VL conditioning space, a projection layer is required for both dimensional and semantic alignment. \sonevl and Qwen2.5-VL use the same tokenizer, visual-token count, and image patch format. Except for the two newly introduced task special tokens, \texttt{<image\_gen>} and \texttt{<image\_edit>}, their text token boundaries are aligned, enabling position-wise hidden-state alignment. Given a hidden state $h_i \in \mathbb{R}^{5120}$ from the backbone, the alignment layer is defined as
\begin{equation}
c_i = W_2 \cdot \mathrm{GELU}(W_1 h_i + b_1) + b_2,
\end{equation}
where
\begin{equation}
W_1 \in \mathbb{R}^{5120 \times 5120}, \quad
W_2 \in \mathbb{R}^{3584 \times 5120}, \quad
c_i \in \mathbb{R}^{3584}.
\end{equation}
Projecting all hidden states yields the diffusion-conditioning sequence:
\begin{equation}
C = \{c_1, c_2, \ldots, c_n\} = f_{\mathrm{align}}(H).
\end{equation}
This layer does not simply feed text embeddings into an image generation model. Instead, it aligns scientific multimodal representations that already fuse text, image content, task intent, and reasoning-process information into the MMDiT conditioning space.

\section{Data Construction}
\label{sec:data}

\subsection{Data Overview}
\label{sec:data_overview}

To support unified training for scientific image understanding, generation, and editing, we construct SciGenEdit, a dataset containing approximately 314K training samples. SciGenEdit combines open-source data and internally synthesized data and covers three major task categories: image generation, image editing, and image understanding. Image generation accounts for 32.80\%, image editing for 60.83\%, and image understanding for 6.37\% of the dataset. The image understanding data mainly comes from the S1-VL training data and is used to preserve scientific multimodal understanding and thinking-with-images capabilities. The image generation and editing data are used to improve scientific image generation, scientific illustration editing, medical image segmentation, image translation, and super-resolution. We strictly exclude benchmark test samples during data construction to avoid train-test contamination and ensure fair evaluation.

\begin{figure}[t!]
    \centering
    \includegraphics[width=0.96\linewidth]{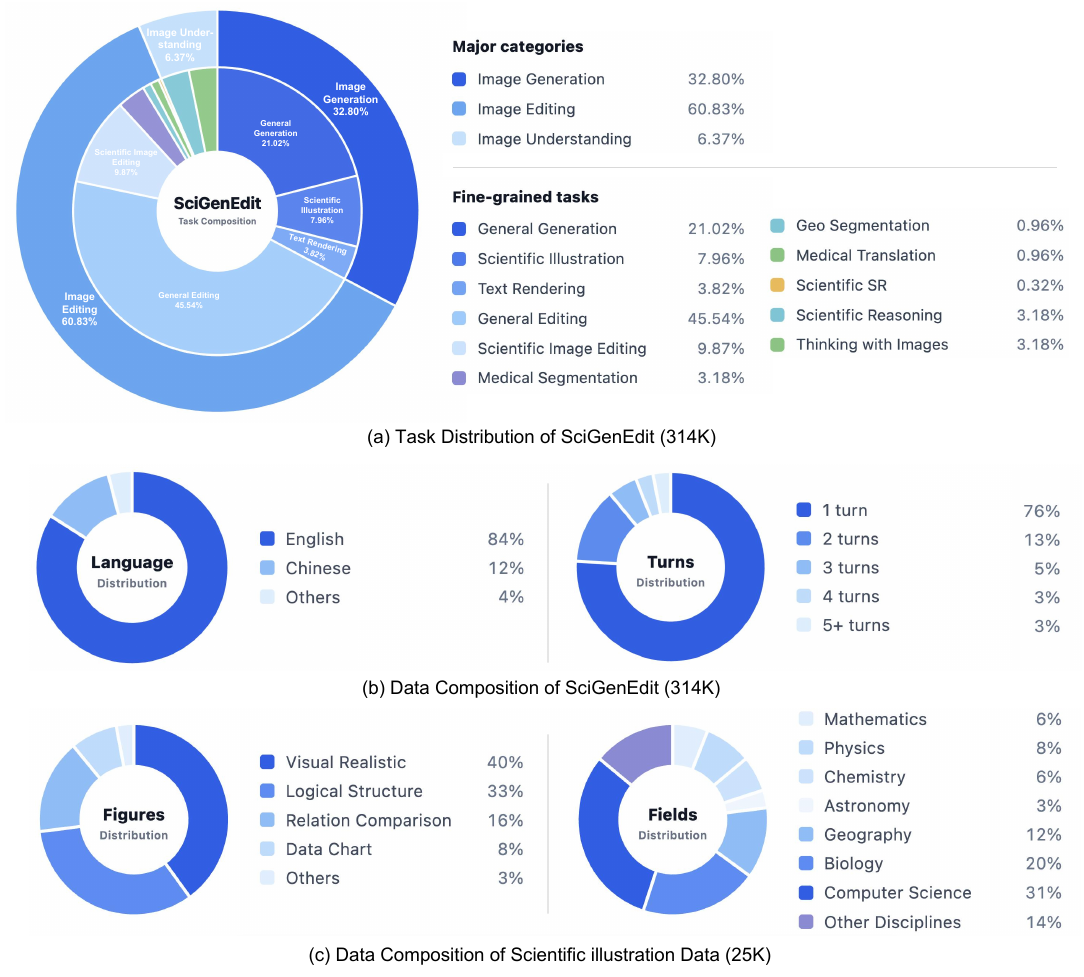}
    \caption{Overview of the SciGenEdit dataset. The figure shows task distribution, language distribution, turn distribution for the full 314K training set, and image-type and discipline distributions for the 25K scientific illustration subset.}
    \label{fig:data}
\end{figure}

As shown in Figure~\ref{fig:data}, SciGenEdit includes a wide range of fine-grained tasks. The image generation portion includes general generation, scientific illustration generation, and text rendering. The image editing portion includes general editing, scientific image editing, medical image segmentation, geographic image segmentation, medical image translation, and scientific image super-resolution. The image understanding portion includes scientific multimodal reasoning and thinking with images. The dataset is primarily English, accounting for 84\% of samples, with Chinese accounting for 12\% and other languages for 4\%. In terms of dialogue turns, single-turn data accounts for 76\%, while multi-turn data covers two or more turns and accounts for 24\%.

We also curate and release a high-quality SciGenEdit-10K subset, focusing on scientific image generation and editing tasks. This subset is constructed through task filtering, quality filtering, and format normalization. It is released together with the model weights and inference service code under the Apache 2.0 license to facilitate open research in scientific image generation and editing.

\subsection{Scientific Image Generation Data}
\label{sec:data_generation}

Scientific image generation data mainly covers scientific illustration generation and text rendering. Scientific illustration generation requires the model to produce scientific images from textual descriptions, paper abstracts, method descriptions, or structured requirements. The resulting images include realistic visualizations, logical diagrams, relational comparisons, data charts, and other scientific image types. Such images commonly contain modules, arrows, labels, hierarchical relations, and scientific symbols, placing strong demands on structural planning, semantic consistency, and text layout.

Within the 25K scientific illustration subset, realistic visualizations account for 40\%, logical diagrams for 33\%, relational comparisons for 16\%, data charts for 8\%, and other types for 3\%. Different illustration types impose different visual requirements: realistic visualizations emphasize object morphology and scientific scenes; logical diagrams focus on module organization, process relations, and system architecture; relational comparisons express differences between concepts, methods, or experimental conditions; and data charts require stronger text, coordinate, and structured-visual expression capability.

The scientific illustration data spans mathematics, physics, chemistry, astronomy, geography, biology, computer science, and other disciplines. This multidisciplinary coverage helps the model learn visual conventions across scientific fields, such as model architecture diagrams and algorithm flowcharts in computer science, mechanism diagrams and experimental workflows in biomedicine, and structural schematics in geography and physics. Text rendering data further improves the model's ability to generate labels, titles, module names, and explanatory text in scientific images.

\subsection{Scientific Image Editing Data}
\label{sec:data_editing}

Scientific image editing data covers general editing, scientific image editing, medical image segmentation, geographic image segmentation, medical image translation, and scientific image super-resolution. Unlike general image editing, scientific image editing must not only execute local visual modifications, but also preserve scientific semantics, structural relations, region localization, and domain constraints.

For scientific illustration editing, we construct triplets consisting of the source image, editing instruction, and target image. Editing instructions include adding or removing modules, modifying labels, adjusting arrow connections, changing layouts, unifying styles, and multi-turn refinement. To simulate realistic scientific illustration workflows, we further construct multi-turn editing data so that the model can progressively modify images according to consecutive user instructions while maintaining global layout and semantic consistency across turns.

For segmentation tasks, we transform conventional mask prediction into a native image editing problem. The model takes the original image and a natural-language instruction as input and outputs an edited image with colored overlays on target regions. Medical images are mostly grayscale, and the overlays use only three high-saturation colors: red, yellow, and green. Their RGB ranges differ clearly from the grayscale background, allowing target regions to be robustly extracted during evaluation and converted back to binary masks. The output image resolution is kept the same as the input resolution. Three-dimensional medical data are processed slice by slice in 2D, and MSD also follows this per-slice setting~\citep{antonelli2022medical}.

For medical image translation and scientific image super-resolution, we use the same unified editing format. SynthRAD2025 is trained and evaluated on 2D slices without explicitly modeling volumetric consistency~\citep{thummerer2025synthrad2025}. IXI covers both 2x and 4x super-resolution settings, with target magnification specified through different prompts~\citep{ixiDataset}. By formulating these tasks as ``input image + editing instruction $\rightarrow$ textual reasoning process + output image'', \model can learn shared structure across different scientific image editing tasks within a single training framework.

\section{Training Strategy}
\label{sec:training}

\begin{figure}[t!]
    \centering
    \includegraphics[width=0.999\linewidth]{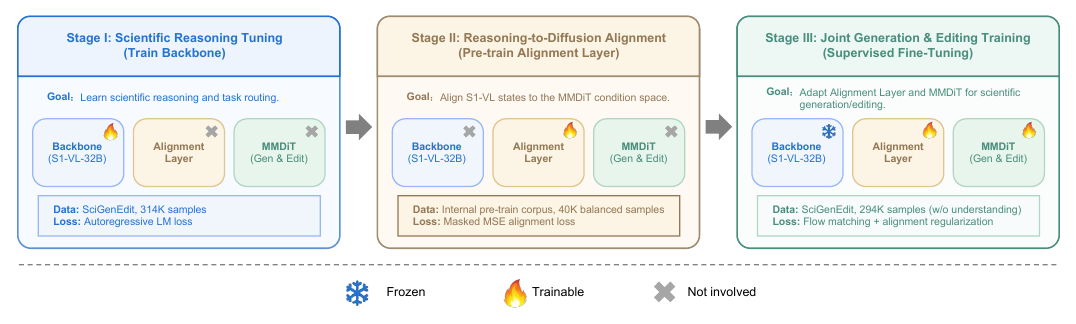}
    \caption{Three-stage training strategy of \model. Stage I performs scientific reasoning tuning, Stage II trains the reasoning-to-diffusion alignment layer, and Stage III jointly optimizes the alignment layer and DiT on SciGenEdit.}
    \label{fig:training}
\end{figure}

\model adopts a three-stage training strategy, as shown in Figure~\ref{fig:training}. In Stage I, we use the full 314K SciGenEdit dataset to perform scientific reasoning tuning on \sonevl, enabling task understanding and reasoning related to scientific image generation and editing. This stage only supervises the front-end textual response, namely explicit reasoning, textual answer, and task special token, and does not require the model to output the final image. In Stage II, we draw 40K samples from an internal pre-training database and train the reasoning-to-diffusion alignment layer from scratch, mapping S1-VL hidden states into the native Qwen-Image MMDiT conditioning space. In Stage III, we use the 294K image generation and editing samples from SciGenEdit, excluding the 20K understanding-only samples, for joint training. The backbone participates in online forward computation but remains frozen, while the alignment layer and MMDiT are trained to enable scientific reasoning representations to drive generation and editing.

\subsection{Stage I: Scientific Reasoning Tuning}
\label{sec:training_stage1}

The goal of Stage I is to enhance \sonevl's reasoning capability for scientific image generation and editing, so that its hidden states can provide high-quality scientific semantic conditions for the diffusion generation module in later stages. Starting from \sonevl, we fully fine-tune the language-model component of the S1-VL-32B vision-language backbone. The training data comes from the full SciGenEdit dataset, containing approximately 314K multimodal dialogue samples. These samples cover scientific illustration understanding and description, image editing intent parsing and reasoning, segmentation target description, medical image task specification, multi-turn scientific illustration editing, and thinking-with-images tasks. Importantly, Stage I does not train the image output itself. For generation and editing samples that include target images, we retain and supervise only the text sequence preceding image generation, including reasoning, textual answer, and task special token.

This stage uses the standard autoregressive language modeling objective. Given input context $x$ (including text instruction and optional image) and target response sequence $y$ (including reasoning, textual answer, and task special token), the objective is
\begin{equation}
\mathcal{L}_{\mathrm{reason}} = -\sum_{t=1}^{T} \log p(y_t \mid y_{<t}, x).
\end{equation}
We use packing and padding-free training to improve GPU utilization. The visual encoder is frozen, only the language model is trained, and sequence parallelism is enabled to support context windows up to 32K tokens.

\subsection{Stage II: Reasoning-to-Diffusion Alignment}
\label{sec:training_stage2}

Stage II aims to bridge the representation gap between the reasoning backbone and the diffusion generation module. \sonevl produces hidden states with dimensionality $d_{\mathrm{src}} = 5120$, whereas Qwen-Image MMDiT natively depends on the $d_{\mathrm{tgt}} = 3584$ conditioning space encoded by Qwen2.5-VL. Therefore, with both backbone models fully frozen, we train the reasoning-to-diffusion alignment layer $\mathcal{A}_{\phi}$ from scratch to learn a mapping from S1-VL hidden states to the Qwen-Image conditioning space. This stage is not intended to inject scientific knowledge by itself. Instead, it serves as condition-space warm-up and manifold initialization, making optimization more stable after replacing the original Qwen2.5-VL condition encoder. The actual adaptation to scientific reasoning conditions is mainly learned in Stage III through diffusion supervision on SciGenEdit. The MLP projector contains approximately 44.57M trainable parameters.

We sample approximately 40K text prompts from an internal pre-training database to construct a balanced corpus covering general and scientific visual descriptions, including object attributes, spatial relations, scene composition, scientific illustration descriptions, and editing instructions. The alignment objective is a masked mean squared error loss:
\begin{equation}
\mathcal{L}_{\mathrm{align}}^{\mathrm{II}} =
\frac{\sum_{i=1}^{B} \sum_{t=1}^{L_i}
\left\| \mathcal{A}_{\phi}(\mathbf{h}^{\mathrm{src}}_{i,t}) -
\mathbf{h}^{\mathrm{tgt}}_{i,t} \right\|_2^2}
{\sum_{i=1}^{B} L_i \cdot d_{\mathrm{tgt}}}.
\end{equation}

\subsection{Stage III: Joint Generation and Editing Training}
\label{sec:training_stage3}

Although Stage II establishes a geometric correspondence between the reasoning and diffusion representation spaces, the alignment projector is trained in isolation from the actual image synthesis pipeline. Stage III eliminates this mismatch by jointly training the alignment projector $\mathcal{A}_{\phi}$ and DiT $\mathcal{G}_{\theta}$ end to end, allowing the two components to co-adapt under a shared diffusion objective while keeping the VLM backbone, VAE, and scheduler frozen.

The joint training dataset contains 294K samples from SciGenEdit, covering text-to-image generation samples $\mathcal{D}_{\mathrm{gen}}$ and image editing samples $\mathcal{D}_{\mathrm{edit}}$, excluding the 20K understanding-only samples with text outputs. A generation sample includes a text prompt and a target image. The model autoregressively generates reasoning, textual answer, and \texttt{<image\_gen>}, and directly collects the hidden states from that generation process as diffusion conditions. An editing sample includes a source image, an editing instruction, and a target image. The model autoregressively generates reasoning, textual answer, and \texttt{<image\_edit>}, and collects hidden states from the same generation pass that jointly encodes the source image, editing instruction, and generated response. Unlike Stage I, which supervises only text outputs, Stage III additionally computes diffusion supervision for samples with image targets, enabling the alignment layer and DiT to learn the mapping from scientific reasoning representations to target images.

The total training loss combines diffusion loss and alignment regularization:
\begin{equation}
\mathcal{L}_{\mathrm{total}} = \mathcal{L}_{\mathrm{diff}} + \lambda \cdot \mathcal{L}_{\mathrm{align}}.
\end{equation}
The diffusion loss follows standard flow-matching supervised fine-tuning, where $w(t)$ denotes the timestep weighting function:
\begin{equation}
\mathcal{L}_{\mathrm{diff}} =
w(t) \cdot
\left\|
\mathcal{G}_{\theta}(\mathbf{z}_t, t, \mathcal{A}_{\phi}(\mathbf{h}^{\mathrm{src}})) -
\mathbf{v}_{\mathrm{target}}
\right\|_2^2 .
\end{equation}
The alignment loss extends the Stage-II MSE objective with a cosine-similarity term:
\begin{equation}
\mathcal{L}_{\mathrm{align}}^{\mathrm{III}}
=
\mathrm{MSE}(\mathcal{A}_{\phi}(\mathbf{h}^{\mathrm{src}}), \mathbf{h}^{\mathrm{tgt}})
+ \frac{1}{2}
\left(
1 - \frac{1}{L}\sum_{t=1}^{L}
\cos(\mathcal{A}_{\phi}(\mathbf{h}^{\mathrm{src}}_t), \mathbf{h}^{\mathrm{tgt}}_t)
\right).
\end{equation}
Here, $\mathbf{h}^{\mathrm{tgt}}_t$ is produced by the frozen Qwen2.5-VL condition encoder and aligned by token position.

We set $\lambda=0.8$. The MLP is initialized from the best Stage-II checkpoint, and DiT is initialized from Qwen-Image pre-trained weights. The model is trained for 8 epochs. During training, we follow the zero\_cond\_t mechanism introduced by Qwen-Image and apply special handling to timestep modulation in the DiT forward pass. This mechanism is not equivalent to dropping condition embeddings with probability $p$ and should not be interpreted as conventional classifier-free guidance training. During inference, we enable classifier-free guidance with the default $\mathrm{cfg\_scale}=4.0$.

\begin{table}[t!]
\centering
\caption{Training objectives, trainable modules, and major hyperparameters of the three \model training stages.}
\label{tab:training}
\scriptsize
\setlength{\tabcolsep}{0.6pt}
\begin{tabularx}{\linewidth}{p{0.20\linewidth}*{3}{>{\centering\arraybackslash}X}}
\toprule
Item & Stage I & Stage II & Stage III \\
\midrule
Objective & Language Modeling & Masked MSE & Flow Matching + Alignment \\
Trainable Modules & S1-VL-32B (LLM) & Alignment MLP & Alignment MLP + DiT \\
Frozen Modules & ViT & S1-VL-32B, Qwen2.5-VL & S1-VL-32B, Qwen2.5-VL, VAE \\
Trainable Params & $\sim$32B & 44.57M & $\sim$20B \\
Optimizer & AdamW & AdamW & AdamW \\
Learning Rate & $1\times10^{-5}$ & $1\times10^{-4}$ & $4\times10^{-5}$ \\
Global Batch Size & 256 & 128 & 64 \\
Steps & $\sim$3.7K & $\sim$4.7K & $\sim$36.8K \\
Max Sequence Length & 32768 & 32768 & 32768 \\
\bottomrule
\end{tabularx}
\end{table}

\subsection{Infrastructure and Hyperparameters}
\label{sec:training_infra}

\model is trained on 8 servers, each equipped with 8 NVIDIA H200 GPUs. Our training infrastructure follows the implementation style of \sonevl, using \texttt{ms-swift} as the multimodal supervised fine-tuning framework and DeepSpeed for distributed training and memory optimization~\citep{zhao2024swiftascalablelightweightinfrastructure,rasley2020deepspeed}. Stage I uses ZeRO-3 and sequence parallelism to support 32K-token long-context training. Stage II trains only the lightweight alignment MLP. Stage III freezes the VLM backbone and VAE, and trains only the alignment MLP and DiT, allowing the reasoning representation and diffusion generation module to jointly adapt to image generation and editing data.

Table~\ref{tab:training} summarizes the main hyperparameters of the three training stages. Global batch size and steps are estimated from the current data scale and epoch settings: Stage I trains on approximately 314K SciGenEdit dialogue samples, Stage II trains on approximately 40K prompts sampled from the internal pre-training database, and Stage III trains on approximately 294K generation and editing samples from SciGenEdit.

\section{Experiments}
\label{sec:experiments}

\subsection{Evaluation Benchmarks}
\label{sec:benchmarks}

\begin{table}[t!]
\centering
\caption{Evaluation benchmarks and task matrix. Benchmarks are grouped into scientific image generation, editing, and understanding, with their task definitions and primary metrics.}
\label{tab:benchmarks}
\scriptsize
\setlength{\tabcolsep}{3pt}
\renewcommand{\arraystretch}{1.08}
\begin{tabularx}{\linewidth}{>{\raggedright\arraybackslash}p{0.25\linewidth}p{0.13\linewidth}X}
\toprule
Task Type & Benchmark & Description \\
\midrule
\rowcolor{gray!12}\multicolumn{3}{l}{\textbf{Image Generation}} \\
Scientific Illustration Generation & GenExam & A multidisciplinary text-to-image examination benchmark covering mathematics, physics, chemistry, biology, computer science, and related fields. The metric is LLM-as-Judge~\citep{GenExam2025}. \\
Scientific Illustration Generation & TechImage-Bench & A rubric-based benchmark for professional image generation, covering biology, engineering, and general diagrams. The metric is LLM-as-Judge~\citep{Ni2025TechImageBench}. \\
Text Rendering & CVTG-2K & A complex visual text generation benchmark with long text and multiple text regions, designed to evaluate text generation quality inside images. The metric is LLM-as-Judge~\citep{du2025textcrafter}. \\
\midrule
\rowcolor{gray!12}\multicolumn{3}{l}{\textbf{Image Editing}} \\
Medical Segmentation & MSD & The Medical Segmentation Decathlon contains ten heterogeneous medical image segmentation tasks across anatomical structures, modalities, and ROI complexity. The metric is Dice~\citep{antonelli2022medical}. \\
Rock Microstructure Segmentation & cigRockSEM & A scanning electron microscopy benchmark for rock microstructure interpretation, providing SEM images and baseline methods. The metric is F1~\citep{zhang2025benchmark}. \\
Medical Image Translation & SynthRAD2025 & A synthetic CT generation challenge dataset with paired MRI-CT and CBCT-CT cases for radiotherapy-oriented sCT evaluation. The metric is PSNR~\citep{thummerer2025synthrad2025}. \\
Pathology Image Translation & BCI & A paired pathology image translation dataset from breast cancer H\&E to IHC images, associated with HER2 expression grades. The metric is PSNR~\citep{liu2022bci}. \\
Medical Super-Resolution & IXI & A multimodal MR dataset from nearly 600 healthy subjects, including T1, T2, PD, MRA, and DWI sequences. The metric is 2x/4x MRI super-resolution PSNR~\citep{ixiDataset}. \\
\midrule
\rowcolor{gray!12}\multicolumn{3}{l}{\textbf{Image Understanding}} \\
Scientific Multimodal Reasoning & MMMU & A large-scale multidisciplinary multimodal understanding and reasoning benchmark covering expert-level knowledge, chart understanding, and cross-disciplinary problem solving. The metric is accuracy~\citep{yue2024mmmu}. \\
Scientific Multimodal Reasoning & SFE & Scientists' First Exam evaluates scientific cognitive ability of MLLMs, focusing on scientific chart understanding, information extraction, and reasoning. The metric is accuracy~\citep{zhou2025scientists}. \\
Mathematical Visual Reasoning & MathVision & A benchmark for multimodal mathematical reasoning over geometry, functions, charts, and symbolic visual problems. The metric is accuracy~\citep{wang2024measuring}. \\
Thinking-with-Images & HRBench-4K & Constructed by cropping query-relevant regions from original 8K images, this split evaluates fine-grained recognition and visual reasoning in local high-resolution scenes. The metric is accuracy~\citep{hrbench}. \\
Thinking-with-Images & HRBench-8K & Built from full original 8K images, this split evaluates global search and localization in complete high-resolution images. The metric is accuracy~\citep{hrbench}. \\
\bottomrule
\end{tabularx}
\end{table}

As shown in Table~\ref{tab:benchmarks}, we evaluate \model from three perspectives: scientific image generation, scientific image editing, and scientific image understanding. Scientific image generation evaluation includes GenExam, TechImage-Bench, and CVTG-2K. Scientific image editing is evaluated by dataset across MSD, cigRockSEM, SynthRAD2025, BCI, and IXI. Image understanding evaluation includes MMMU, SFE, MathVision, HRBench-4K, and HRBench-8K, and is used to verify whether the model preserves the backbone's understanding capability after introducing generation and editing modules.

For segmentation-as-editing tasks, the model outputs a colored overlay image with the same resolution as the input. During evaluation, target regions are extracted from the red, yellow, and green overlay colors according to their RGB ranges, converted into binary masks, and compared with ground-truth binary masks using mIoU, Dice, or F1. Three-dimensional medical images, including MSD, are processed slice by slice in 2D. cigRockSEM evaluates 2D segmentation quality for rock microstructures. SynthRAD2025 evaluates MRI/CBCT-to-CT translation quality on 2D slices without computing volumetric consistency. BCI evaluates structure preservation and color generation for H\&E-to-IHC pathology translation. IXI evaluates both 2x and 4x MRI super-resolution using PSNR. Image understanding evaluation uses the text branch of \model and does not pass through the diffusion module.

\subsection{Baselines}
\label{sec:baselines}

\paragraph{Scientific image generation.}
To evaluate scientific image generation, we compare \model with representative open-source and closed-source image generation models. Qwen-Image~\citep{Wu2025QwenImage} is used as the open-source foundation baseline because \model reuses its VAE and MMDiT initialization. GLM-Image~\citep{glmimage2026} is included as another strong open-source generation/editing model. We further compare with leading closed-source systems, including Nano Banana 2 (Gemini-3.1-flash-image-preview)~\citep{raisinghani2026nano} and GPT-Image-2~\citep{chatgptimages2026}, to measure the remaining gap to frontier proprietary models.

\paragraph{Scientific image editing.}
For scientific image editing, we evaluate both general-purpose image editing models and task-specific scientific models. The general-purpose baselines include Qwen-Image, GLM-Image, Nano Banana 2, and GPT-Image-2. In addition, because scientific editing benchmarks often have specialized solutions, we compare with task-specific models for each dataset: Swin UNETR~\citep{tang2022self} for MSD, the U-Net baseline from cigRockSEM~\citep{zhang2025benchmark}, VBoussot~\citep{boussot2025registration} for SynthRAD2025, Pyramid Pix2pix~\citep{liu2022bci} for BCI, and CodeBrain~\citep{wu2026virtual} for IXI.

\paragraph{Scientific image understanding.}
For scientific image understanding, we follow the evaluation setting of S1-VL and compare with strong general and scientific image understanding models, including Qwen3-VL-32B-Thinking, Qwen3-VL-235B-A22B-Thinking~\citep{bai2025qwen3}, GPT-5~\citep{singh2025openai}, S1-VL-32B-SFT, S1-VL-32B-RL, and Thyme-VL~\citep{zhang2025thyme}. These comparisons are used to verify whether adding image generation and editing capability preserves the original scientific understanding ability of the S1-VL backbone.

\subsection{Main Results}
\label{sec:results}

\subsubsection{Scientific Image Generation}
\label{sec:results_generation}

\begin{table}[t!]
\centering
\caption{Results on scientific image generation benchmarks.}
\label{tab:generation_results}
\scriptsize
\setlength{\tabcolsep}{4pt}
\begin{tabularx}{\linewidth}{p{0.30\linewidth}*{3}{>{\centering\arraybackslash}X}}
\toprule
Model & GenExam $\uparrow$ & TechImage-Bench $\uparrow$ & CVTG-2K $\uparrow$ \\
\midrule
Qwen-Image & 23.8 & 51.11 & 72.38 \\
GLM-Image & 46.1 & 54.47 & \textbf{91.16} \\
GPT-Image-2 & \textbf{93.8} & \textbf{83.87} & 77.94 \\
Nano Banana 2 & \underline{92.6} & \underline{78.38} & 77.88 \\
\textbf{\model{}} & 53.2 & 61.00 & \underline{80.59} \\
\bottomrule
\end{tabularx}
\end{table}

Table~\ref{tab:generation_results} summarizes the performance on scientific image generation benchmarks. Compared with the Qwen-Image foundation model from which our image module is initialized, \model substantially improves GenExam from 23.8 to 53.2 and TechImage-Bench from 51.11 to 61.00. It also outperforms GLM-Image on both scientific illustration benchmarks, indicating that the proposed scientific reasoning supervision and reasoning-conditioned generation are particularly effective for structure-heavy scientific diagrams. On CVTG-2K, \model reaches 80.59, improving over Qwen-Image and ranking second overall, although GLM-Image remains strongest on this text-rendering benchmark.

Compared with frontier closed-source systems, \model still has a clear gap on GenExam and TechImage-Bench. GPT-Image-2 and Nano Banana 2 achieve 93.8/83.87 and 92.6/78.38, respectively, reflecting the advantage of large-scale proprietary pre-training in visual aesthetics, prompt following, and dense layout synthesis. Nevertheless, among open-weight models, \model provides a substantially stronger scientific generation baseline, especially on benchmarks that require scientific structure, relational consistency, and diagrammatic organization rather than only visual appeal.

\begin{figure}[p]
    \centering
    \includegraphics[width=\textwidth,height=1.0\textheight,keepaspectratio]{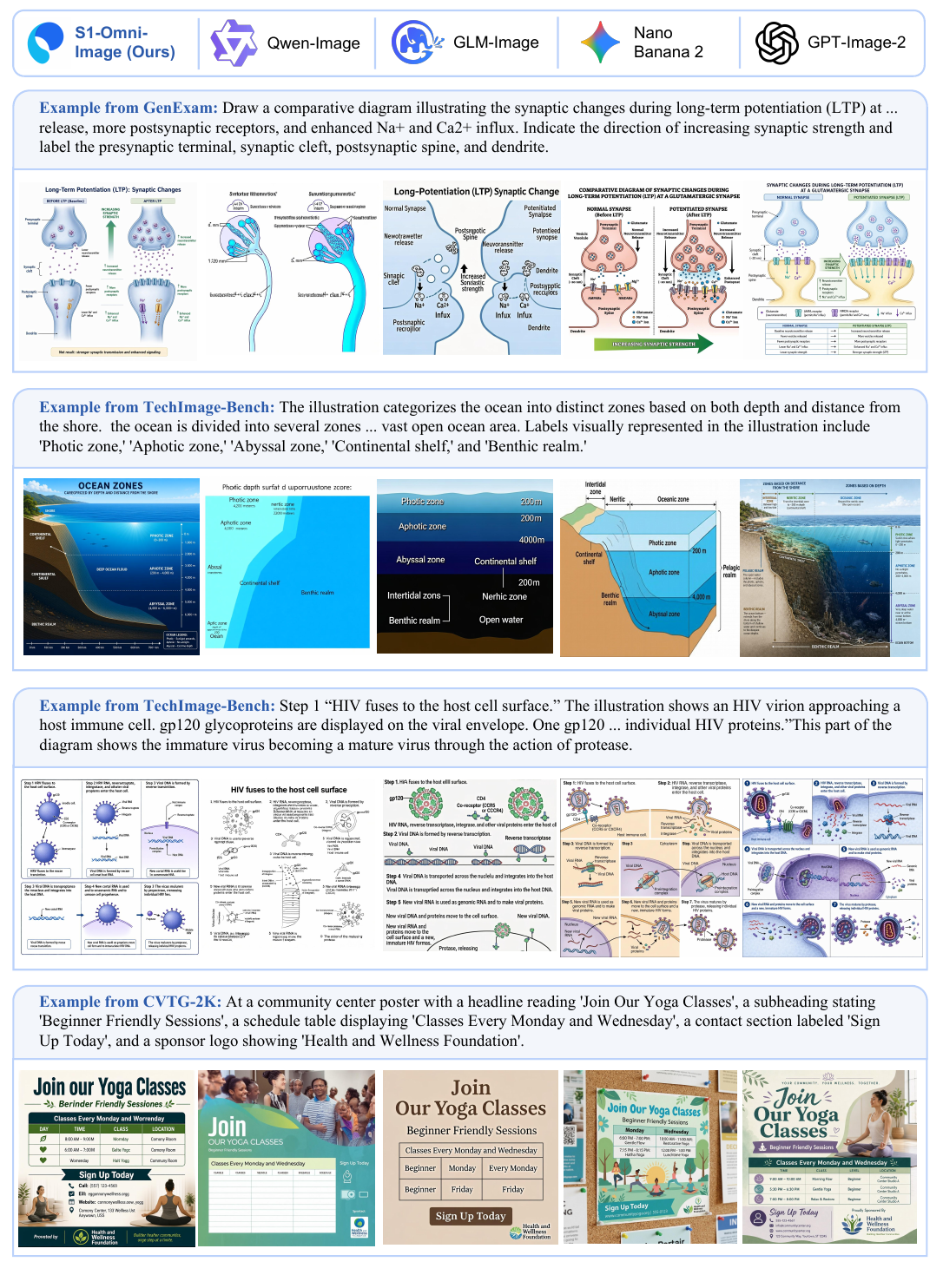}
    \caption{Qualitative comparison with representative models on scientific illustration generation benchmarks. The examples cover GenExam, TechImage-Bench, and CVTG-2K, evaluating scientific illustration generation and complex text rendering.}
    \label{fig:gen_bench}
\end{figure}

The qualitative comparisons in Figure~\ref{fig:gen_bench} provide more direct evidence for this behavior. In the GenExam case on long-term potentiation (LTP), the prompt asks for a comparative synaptic diagram that shows increased neurotransmitter release, more postsynaptic receptors, and enhanced Na$^+$ and Ca$^{2+}$ influx. Qwen-Image produces blurred and unstable text, and the left-right comparison does not faithfully express the requested biological contrast. GLM-Image also fails to produce a clean scientific diagram: the figure is visually cluttered, the color palette is monotonous, and the result lacks the explanatory style expected in scientific illustrations. In contrast, \model produces a clearer comparative diagram with explicit presynaptic and postsynaptic structures, visible neurotransmitter release, receptor changes, ion influx, and an overall layout that is closer in style to the closed-source models.

The remaining examples in Figure~\ref{fig:gen_bench} show similar trends. For ocean-zone illustration and HIV-related scientific diagrams, \model better preserves the high-level scientific layout and diagrammatic organization than open-source baselines, while closed-source models still tend to produce more polished visual designs. For CVTG-2K, the model can render structured poster content better than Qwen-Image, but local text errors and dense-label instability remain. Overall, these results suggest that \model narrows the open-source gap in scientific illustration generation by using explicit reasoning as a semantic and structural plan before image synthesis, while further improvements in text rendering and visual refinement remain important future directions.

\subsubsection{Scientific Image Editing}
\label{sec:results_editing}

\begin{table}[t!]
\centering
\caption{Results on scientific image editing benchmarks.}
\label{tab:editing_results}
\scriptsize
\setlength{\tabcolsep}{2pt}
\begin{tabularx}{\linewidth}{p{0.24\linewidth}*{5}{>{\centering\arraybackslash}X}}
\toprule
Model & MSD $\uparrow$ & cigRockSEM $\uparrow$ & SynthRAD2025 $\uparrow$ & BCI $\uparrow$ & IXI $\uparrow$ \\
\midrule
Task-Specific Models & \underline{0.7868} & \underline{0.9733} & 34.78 & \textbf{21.16} & 24.67 \\
Qwen-Image & 0.0396 & 0.9210 & \underline{42.03} & 0 & 14.25 \\
GLM-Image & 0.0442 & 0.6394 & 35.00 & 12.74 & 11.66 \\
GPT-Image-2 & 0.3787 & 0.9001 & 40.49 & 14.30 & 19.96 \\
Nano Banana 2 & 0.3263 & 0.8962 & 40.08 & 14.33 & \underline{25.27} \\
\textbf{\model{}} & \textbf{0.8528} & \textbf{0.9824} & \textbf{47.92} & \underline{20.67} & \textbf{26.79} \\
\bottomrule
\end{tabularx}
\end{table}

Table~\ref{tab:editing_results} shows that \model achieves the best performance on four of five scientific image editing benchmarks: MSD, cigRockSEM, SynthRAD2025, and IXI. On MSD, \model reaches 0.8528 Dice, surpassing both general-purpose image editing models and the task-specific Swin UNETR baseline. On cigRockSEM, \model achieves 0.9824 F1, exceeding the dedicated U-Net benchmark baseline as well as all general image models. On SynthRAD2025 and IXI, \model also obtains the highest PSNR, reaching 47.92 and 26.79, respectively. These results indicate that the unified editing formulation is effective not only for segmentation-as-editing, but also for translation-as-editing and super-resolution-as-editing.

Compared with general-purpose models, the advantage is especially clear on scientific segmentation tasks. Qwen-Image, GLM-Image, GPT-Image-2, and Nano Banana 2 achieve much lower scores on MSD, showing that generic image editing models struggle to convert precise scientific instructions into spatially accurate overlays. Even strong closed-source models remain far below \model on MSD and cigRockSEM, suggesting that scientific editing requires domain-aware localization and preservation constraints rather than generic visual manipulation. Compared with task-specific models, \model is also competitive: it surpasses the specialist baselines on MSD, cigRockSEM, SynthRAD2025, and IXI, while remaining second on BCI, where Pyramid Pix2pix still benefits from task-specific pathology stain translation.

\begin{figure}[p]
    \centering
    \includegraphics[width=1.0\textwidth,height=1.0\textheight,keepaspectratio]{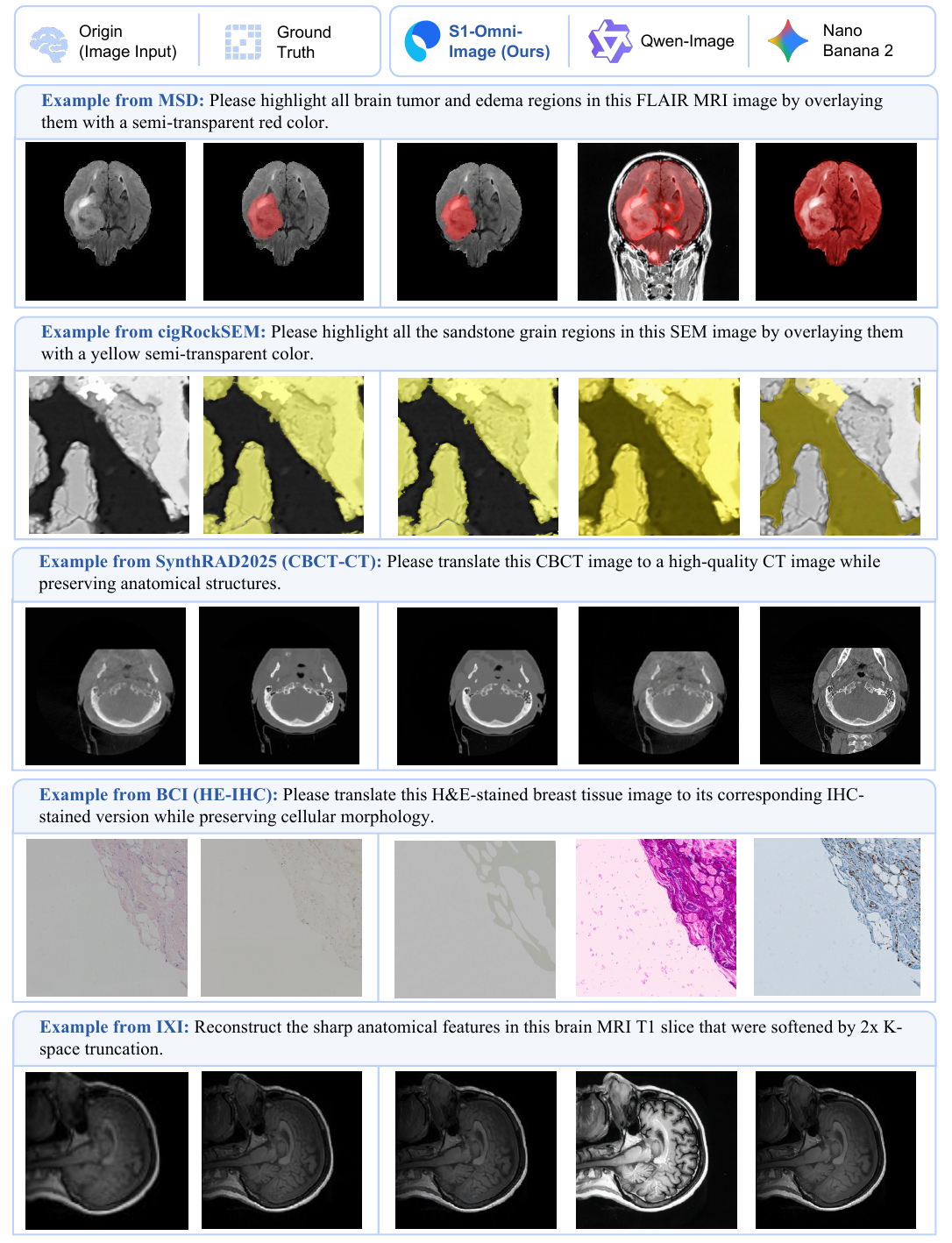}
    \caption{Qualitative comparison with representative models on scientific image editing benchmarks. The examples cover medical image segmentation, rock microstructure segmentation, medical image translation, pathology image translation, and medical super-resolution.}
    \label{fig:edit_bench}
\end{figure}

The qualitative cases in Figure~\ref{fig:edit_bench} further explain these results. In the MSD example, the instruction asks the model to highlight brain tumor and edema regions in a FLAIR MRI slice. \model produces a localized semi-transparent red overlay that closely follows the abnormal region and preserves the surrounding anatomy. By contrast, general-purpose baselines fail to treat the task as precise medical segmentation: Qwen-Image introduces substantial anatomical distortion, while Nano Banana 2 overpaints large portions of the brain rather than isolating the target region. This illustrates why high-level image editing ability alone is insufficient for scientific segmentation, where the model must identify the target structure and preserve the diagnostic context.

The cigRockSEM case shows a similar pattern in a non-medical scientific domain. \model highlights the sandstone grain region with a spatially coherent yellow overlay that aligns with the ground-truth microstructure. Qwen-Image partially captures the target but produces less precise boundaries, while Nano Banana 2 assigns the overlay to an incorrect region and fails to preserve the intended segmentation semantics. Across the SynthRAD2025, BCI, and IXI cases, \model also better maintains anatomical or tissue structure during translation and restoration. These observations support our central claim: by first reasoning about the task type, target region, and preservation constraints, \model can turn diverse scientific editing problems into instruction-conditioned native image editing while retaining domain-specific spatial fidelity.

\clearpage
\clearpage

\clearpage
\clearpage

\subsubsection{Scientific Image Understanding}
\label{sec:results_understanding}

\begin{table}[t!]
\centering
\caption{Results on scientific image understanding and thinking-with-images benchmarks.}
\label{tab:understanding_results}
\scriptsize
\setlength{\tabcolsep}{0.6pt}
\begin{tabularx}{\linewidth}{p{0.33\linewidth}*{5}{>{\centering\arraybackslash}X}}
\toprule
Model & MMMU $\uparrow$ & SFE $\uparrow$ & MathVision $\uparrow$ & HRBench-4K $\uparrow$ & HRBench-8K $\uparrow$ \\
\midrule
Thyme-VL & 48.66 & 25.37 & 26.28 & 77.00 & 72.00 \\
Qwen3-VL-32B-Thinking & 76.00 & 37.50 & 71.51 & 82.63 & 77.00 \\
Qwen3-VL-235B-A22B-Thinking & 77.89 & 39.98 & 74.87 & 83.00 & 80.40 \\
GPT-5 & 81.22 & \textbf{44.06} & 75.66 & 74.25 & 73.75 \\
S1-VL-32B-SFT & 82.50 & 42.58 & 75.89 & 85.00 & \underline{85.10} \\
S1-VL-32B-RL & \textbf{83.40} & \underline{43.10} & \textbf{77.70} & \textbf{91.38} & \textbf{93.50} \\
\textbf{\model{}} & \underline{82.67} & 42.32 & \underline{76.71} & \underline{88.38} & 85.00 \\
\bottomrule
\end{tabularx}
\end{table}

Table~\ref{tab:understanding_results} evaluates whether \model preserves the scientific image understanding capability of the S1-VL backbone after adding image generation and editing. On scientific reasoning benchmarks, \model remains highly competitive. It achieves 82.67 on MMMU and 76.71 on MathVision, ranking second only to S1-VL-32B-RL and outperforming both Qwen3-VL-32B-Thinking and Qwen3-VL-235B-A22B-Thinking. On SFE, \model reaches 42.32, which is close to S1-VL-32B-SFT and S1-VL-32B-RL, and remains far above the Qwen3-VL baselines. These results indicate that the scientific reasoning ability of the backbone is largely preserved despite the addition of image generation and editing modules.

For thinking-with-images benchmarks, \model also maintains strong performance. On HRBench-4K, it achieves 88.38, outperforming Thyme-VL, Qwen3-VL-32B-Thinking, Qwen3-VL-235B-A22B-Thinking, GPT-5, and S1-VL-32B-SFT, and remains close to S1-VL-32B-RL. On HRBench-8K, \model obtains 85.00, matching the strong SFT backbone level but still trailing the RL-enhanced S1-VL-32B-RL. This gap is expected because S1-VL-32B-RL is further optimized for long-horizon high-resolution visual search through reinforcement learning, whereas \model focuses on unifying understanding, generation, and editing under supervised training.

\begin{figure}[t]
    \centering
    \includegraphics[width=1.0\textwidth,height=1.0\textheight,keepaspectratio]{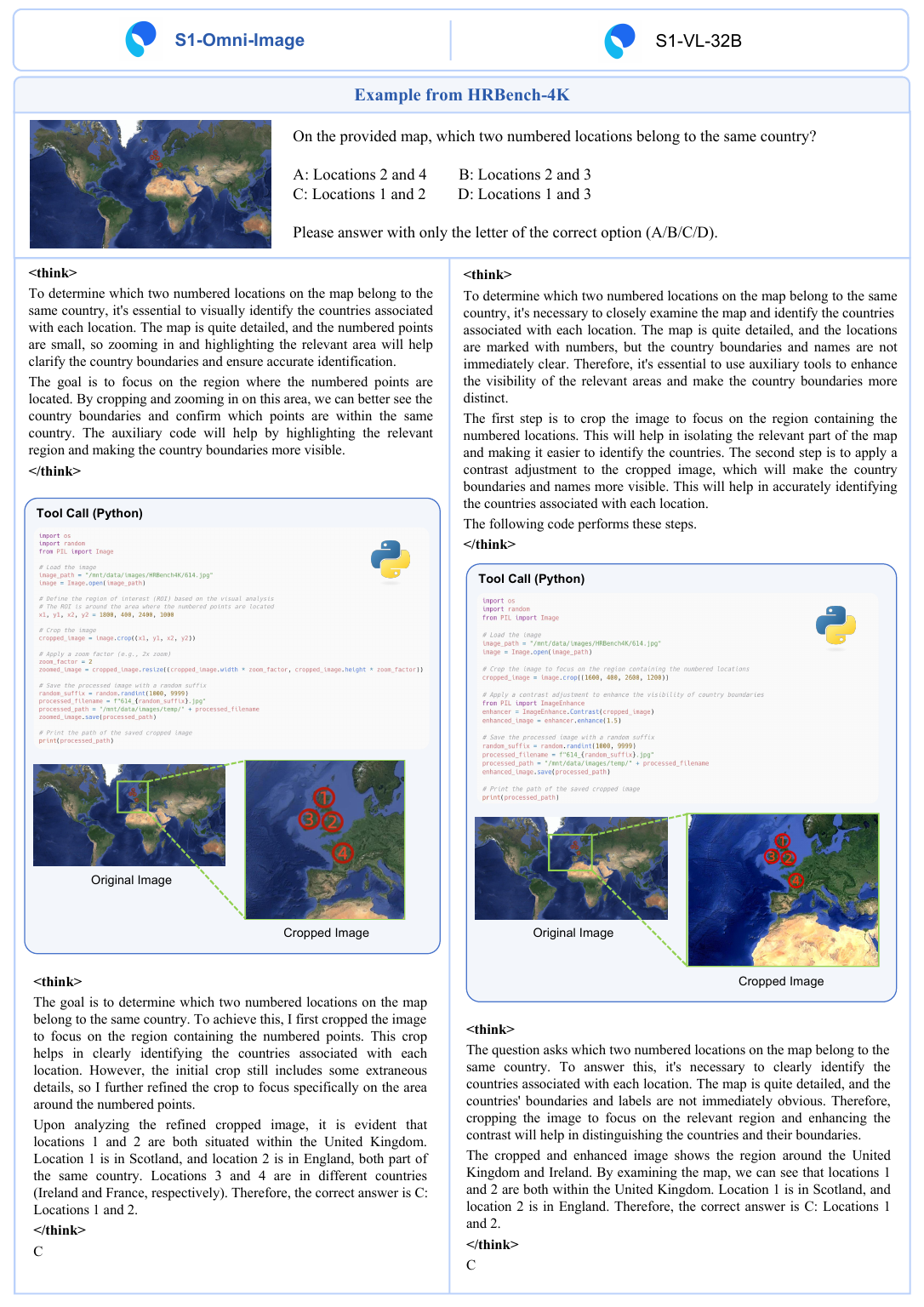}
    \caption{Qualitative comparison on image understanding tasks. The example is from HRBench-4K in the Thinking-with-Images benchmark, comparing our model with the multimodal foundation model S1-VL-32B on high-resolution visual reasoning and understanding.}
    \label{fig:understanding_bench}
\end{figure}

Figure~\ref{fig:understanding_bench} provides a qualitative example from HRBench-4K, where the model must reason over a low-resolution global map and identify the relevant geographic region before answering. Both \model and S1-VL-32B follow the thinking-with-images paradigm by invoking Python-based visual operations during reasoning. S1-VL-32B performs not only region cropping, but also contrast enhancement on the relatively dark map, making the cropped region visually clearer. In comparison, \model uses a simpler operation sequence and mainly performs cropping, but its selected crop is more tightly focused on the task-relevant area. This more concentrated visual evidence is sufficient for the model to identify the correct region and produce the correct answer. The case suggests that \model retains the core multimodal reasoning behavior of S1-VL-32B: it can actively localize useful visual evidence, manipulate the image through tool calls, and complete the reasoning process based on the resulting intermediate observation, even after being extended with image generation and editing capabilities.

Overall, the understanding results demonstrate a favorable trade-off. \model is not intended to surpass the RL-specialized S1-VL-32B-RL on every reasoning benchmark; instead, it preserves most of the backbone's scientific reasoning and thinking-with-images capability while extending the model to scientific image generation and editing.

\section{Analysis}
\label{sec:analysis}

\subsection{Effect of Think-Before-Generate}
\label{sec:analysis_tbg}

\begin{figure}[!t]
    \centering
    \includegraphics[width=\textwidth,height=\textheight,keepaspectratio]{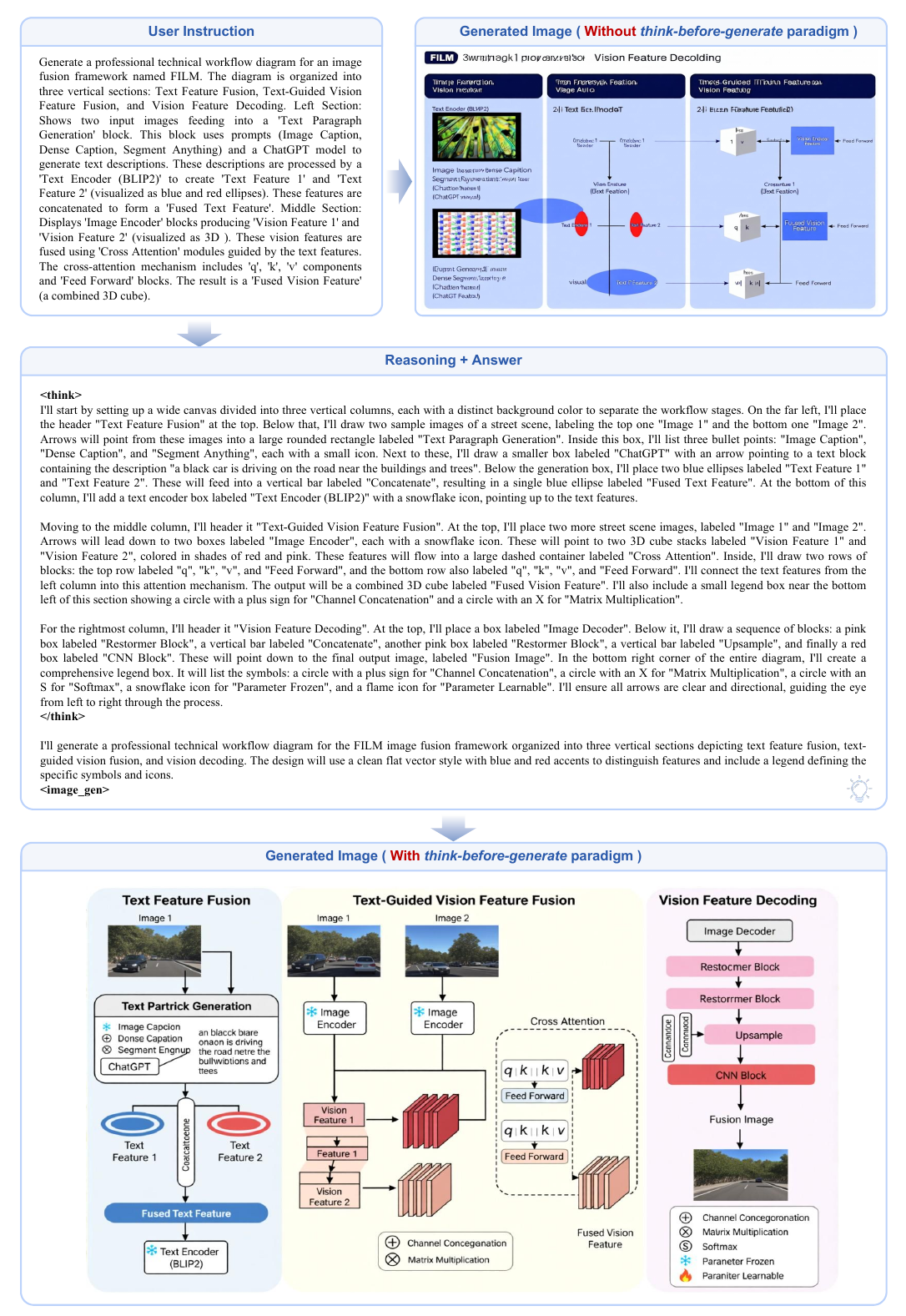}
    \caption{Think-before-generate case in scientific illustration generation. The model first produces task-relevant reasoning and layout planning, and then generates a structured scientific illustration conditioned on the reasoning representation.}
    \label{fig:case_gen}
\end{figure}

Figure~\ref{fig:case_gen} analyzes the effect of the proposed think-before-generate paradigm through a representative scientific illustration generation case. The upper branch shows the result produced without the think-before-generate paradigm, where the model directly maps the user instruction to an image. Although the generated image contains several visually relevant components, it fails to reliably organize the scientific content into a coherent explanatory structure. The layout is fragmented, the module hierarchy is unclear, and several visual elements are only weakly connected to the intended scientific workflow. This suggests that direct prompt-to-image generation is insufficient for structure-heavy scientific illustrations, where the model must first identify key concepts, decide their roles, and arrange them into a logically consistent diagram.

In contrast, the lower branch in Figure~\ref{fig:case_gen} shows the result after introducing think-before-generate training. Before image synthesis, the model produces an explicit reasoning-and-answer sequence that decomposes the instruction into scientific modules, clarifies the relationship between text features and vision features, and plans the visual organization of the final figure. This intermediate reasoning acts as a semantic and structural plan. Conditioned on this reasoning representation, the generated image exhibits clearer module separation, more faithful label assignment, and more coherent flow from input features to fusion and decoding. The resulting illustration better follows scientific visual conventions, such as left-to-right information flow, grouped functional blocks, arrows indicating data transformation, and labels that correspond to the underlying method components.

This comparison supports the central motivation of \model: scientific image generation should not be treated as pure visual synthesis, but as reasoning-conditioned visual construction. Scientific figures often encode abstract concepts, algorithmic procedures, causal relations, and multi-stage pipelines. If the model generates images without first forming an explicit task plan, small semantic errors in early visual elements can propagate into global structural inconsistency. The think-before-generate paradigm mitigates this issue by forcing the model to first construct a textual reasoning trace and then inject the corresponding hidden states into the diffusion module. As a result, the image decoder receives not only the original prompt semantics, but also a task-oriented representation of the intended diagram structure.

The benefit is especially evident in scientific illustration generation because correctness depends on both local details and global organization. The case in Figure~\ref{fig:case_gen} shows that the reasoning process improves three aspects simultaneously: semantic grounding, by identifying the core scientific entities to be visualized; structural planning, by determining how modules should be arranged and connected; and visual execution, by producing a cleaner and more interpretable scientific diagram. These observations are consistent with the quantitative improvements on GenExam and TechImage-Bench, where \model outperforms open-source baselines on scientific structure expression and instruction consistency.

\subsection{Cross-Task Reasoning Cases}
\label{sec:analysis_cross_task}

\begin{figure}[t!]
    \centering
    \includegraphics[width=\textwidth,height=0.9\textheight,keepaspectratio]{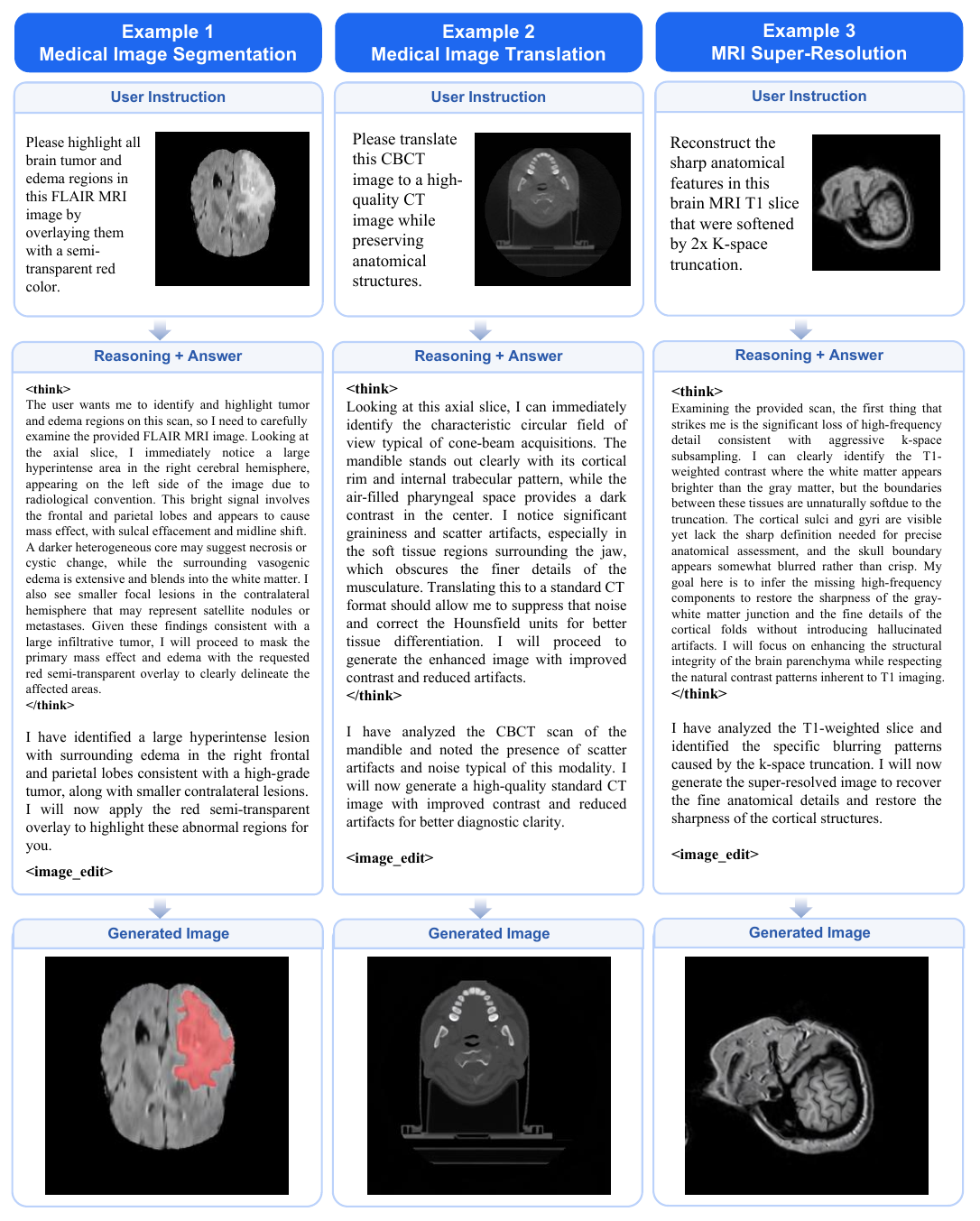}
    \caption{Medical segmentation, image translation, and MRI super-resolution as image editing tasks. It uses reasoning to guide region localization, modality transformation, and detail restoration within a unified framework.}
    \label{fig:case_seg}
\end{figure}

Figure~\ref{fig:case_seg} further demonstrates that think-before-generate is not limited to scientific illustration generation, but can also provide a unified reasoning interface for heterogeneous scientific image editing tasks. The three examples cover medical image segmentation, medical image translation, and MRI super-resolution. Although these tasks differ substantially in their output requirements, they share the same input-output protocol in \model: the user provides an image and a natural-language instruction, the model first generates task-specific reasoning and an answer, and the aligned reasoning hidden states then condition the image editing module to produce the final image.

For medical image segmentation, the reasoning process focuses on identifying the target anatomical or pathological region, interpreting the instruction, and deciding how the target should be represented in the output image. Instead of producing a separate mask tensor, \model casts segmentation as an editing task and generates a colored overlay on the original image. This formulation preserves the natural-language interaction format while still allowing the output to be converted back into a binary mask for quantitative evaluation. The case shows that the model can use its reasoning trace to localize the target region and generate a visually interpretable segmentation result.

For medical image translation, the reasoning process changes accordingly. The model must understand that the task is not to highlight a region or modify the layout, but to transform the image appearance from one modality to another while preserving anatomical structure. In the example, the model reasons about the modality conversion objective and then generates an edited image that changes imaging contrast and texture while maintaining the spatial organization of the input. This indicates that the same reasoning-to-diffusion pathway can support cross-modal image transformation, rather than only local editing or region annotation.

For MRI super-resolution, the model again adopts a different reasoning behavior. The reasoning trace emphasizes recovering fine-grained anatomical details and improving image sharpness while avoiding semantic changes. The generated result enhances structural clarity without altering the main anatomy. This case illustrates that \model can interpret super-resolution as an instruction-conditioned editing task and use reasoning to constrain the diffusion module toward detail restoration rather than content modification.

These cross-task examples show that the proposed paradigm provides a common abstraction for scientific image editing. Segmentation requires region-level localization, translation requires modality-level appearance transformation, and super-resolution requires detail-level reconstruction, yet all can be handled through the same sequence: understand the instruction, reason about the task objective, emit the task router token, and generate the edited image conditioned on reasoning representations. This unified design reduces the need for separate task-specific architectures and helps explain why \model performs competitively across diverse scientific editing benchmarks.

\clearpage

\section{Case Study}
\label{sec:case}

\subsection{Scientific Image Generation}
\label{sec:case_generation}

\begin{figure}[p]
    \centering
    \includegraphics[width=\textwidth,height=1.0\textheight,keepaspectratio]{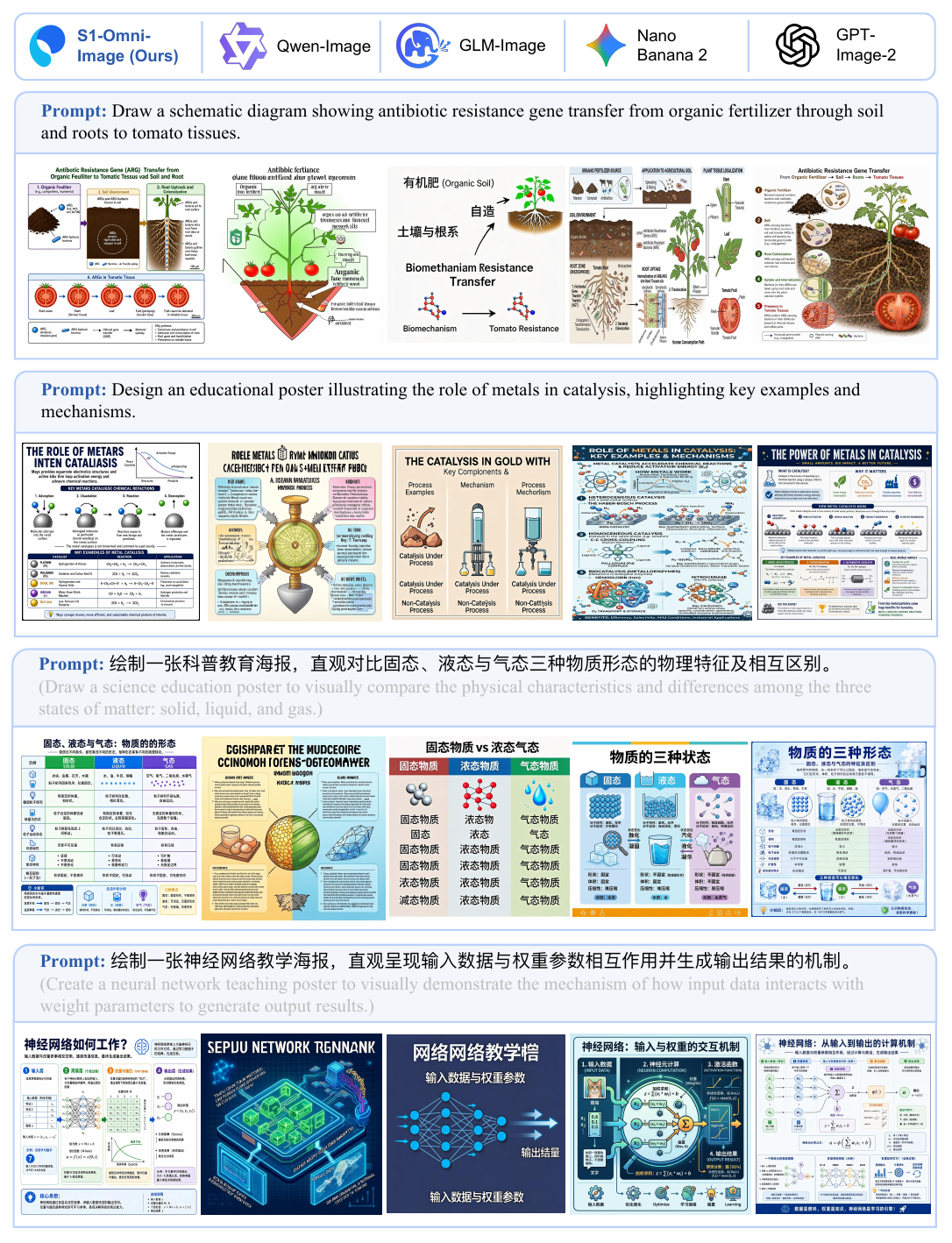}
    \caption{Qualitative comparison with representative image generation models. Under the same prompts, \model more consistently preserves scientific structures, arrow relations, label semantics, and global logical coherence. Compared with other open-source models, it shows clear advantages in scientific visual style and approaches closed-source model behavior in several structure-heavy scientific cases.}
    \label{fig:case_compare}
\end{figure}

Figure~\ref{fig:case_compare} presents qualitative comparisons on real-world scientific illustration prompts in both English and Chinese. The examples cover a diverse set of disciplines and illustration types, including a biology/agricultural mechanism diagram about antibiotic resistance gene transfer, a chemistry educational poster on metal catalysis, a Chinese physics education poster comparing solids, liquids, and gases, and a Chinese computer science teaching poster explaining neural network computation. This diversity is important because scientific illustration generation requires more than visual realism: the model must also organize concepts, preserve domain-specific relations, render labels, and follow the visual conventions of different scientific fields.

Across these cases, open-source baselines often struggle with scientific structure and readability. Qwen-Image can generate visually relevant elements, but its layouts are frequently less organized and its labels may be unstable or semantically incomplete. GLM-Image sometimes produces visually dense posters, yet the scientific organization can be cluttered, with weaker hierarchy and less consistent diagrammatic style. In contrast, closed-source flagship models such as Nano Banana 2 and GPT-Image-2 generally produce polished layouts and stronger overall aesthetics. However, \model produces results that are much closer to these flagship systems in scientific visual style while remaining open-weight: it better preserves module relations, explanatory flow, and the intended scientific semantics.

The first case in Figure~\ref{fig:case_compare} illustrates this behavior clearly. The prompt asks for a schematic diagram showing antibiotic resistance gene transfer from organic fertilizer through soil and roots to tomato tissues. \model organizes the process into a coherent scientific pathway, explicitly showing the source, soil/root transport, transfer mechanism, and tomato-related outcome. Compared with open-source baselines, the generated figure better follows the style of a scientific mechanism diagram and avoids reducing the task to a generic plant or soil illustration. Similar patterns appear in the chemistry and physics poster cases: \model more consistently structures the visual content into interpretable panels, keeps the comparison logic visible, and produces figures that better match educational and scientific communication scenarios.

The Chinese cases further show that \model can handle bilingual scientific illustration prompts. For the physics poster comparing the three states of matter, the model needs to express categorical contrasts and physical properties in a structured educational layout. For the neural network teaching poster, it must represent inputs, weights, computation, and output mechanisms in a diagrammatic form. These cases suggest that the advantage of \model comes from reasoning-conditioned generation: before image synthesis, the model forms an intermediate semantic plan over scientific entities, relations, and layout requirements, and the resulting reasoning representation guides the image decoder toward more structured and scientifically meaningful outputs.

\subsection{Scientific Image Editing}
\label{sec:case_editing}

\begin{figure}[p]
    \centering
    \includegraphics[width=\textwidth,height=1.0\textheight,keepaspectratio]{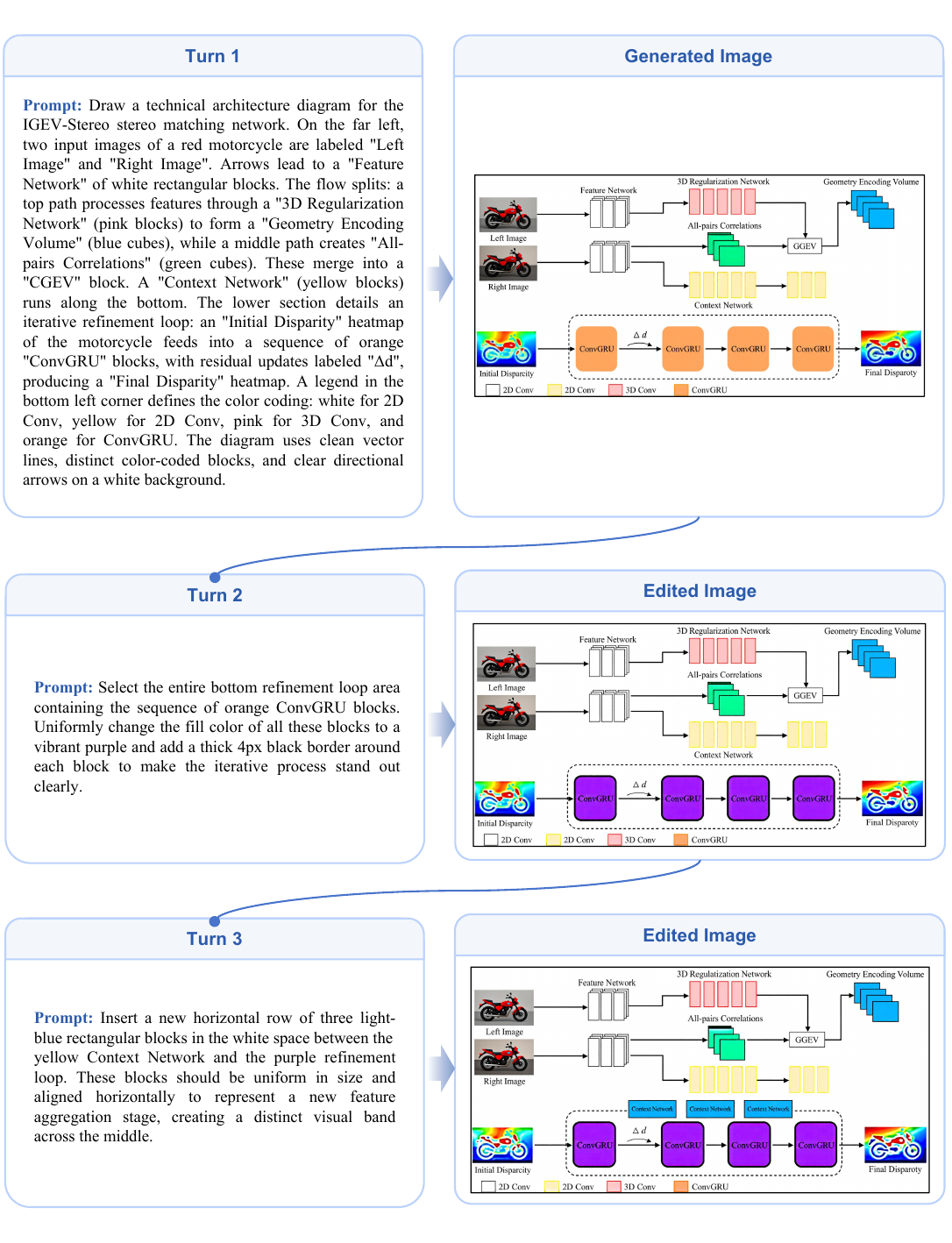}
    \caption{Multi-turn scientific illustration editing case. The model first generates an initial scientific illustration from the user prompt, and then progressively edits module colors and details, adds modules and text, and preserves cross-turn structure and layout consistency according to consecutive editing instructions. For space, reasoning traces are omitted and only the input prompt, editing instructions, and image outputs are shown.}
    \label{fig:case_multiturn}
\end{figure}

Figure~\ref{fig:case_multiturn} demonstrates a multi-turn scientific illustration editing case. We choose a realistic model-architecture diagram from stereo matching as the initial generation target. The first turn asks the model to generate an IGEV-Stereo-style architecture diagram, including left and right input images, feature extraction, 3D regularization, context modeling, ConvGRU-based iterative refinement, disparity maps, arrows, and a legend. The generated image establishes the main visual structure of the scientific diagram, including the upper feature-processing pipeline and the lower iterative refinement loop.

The second turn evaluates local editing under structure preservation. The user asks the model to select the bottom refinement loop containing the orange ConvGRU blocks, change these blocks to purple, and add thick black borders around them. \model performs the requested color and style modification while preserving the rest of the diagram, including the input images, feature-network blocks, arrows, context network, GGEV block, disparity maps, and legend. This shows that the model can localize the target region described by language and apply a visual edit without unnecessarily changing unrelated scientific content.

The third turn further tests multi-turn consistency and additive editing. The user asks the model to insert a new horizontal row of three light-blue rectangular blocks in the white space between the yellow Context Network and the purple refinement loop, with text labels indicating a new feature aggregation stage. \model adds the requested modules at the correct location while maintaining the previous purple ConvGRU edits from Turn 2. The original diagram structure, arrow flow, and surrounding components remain largely stable. This is important for scientific illustration workflows, where users often iteratively refine a figure by changing colors, adding modules, adjusting labels, or reorganizing local structures.

Overall, this case shows that \model supports not only one-shot scientific illustration generation, but also iterative figure editing. The model can maintain visual context across turns, preserve previous edits, and incorporate new requirements without rebuilding the image from scratch. This multi-turn capability is useful for real scientific figure creation, where diagrams are usually refined through a sequence of small, localized modifications rather than completed in a single generation step.

\clearpage

\clearpage

\clearpage

\begin{figure}[t!]
    \centering
    \includegraphics[width=0.96\linewidth]{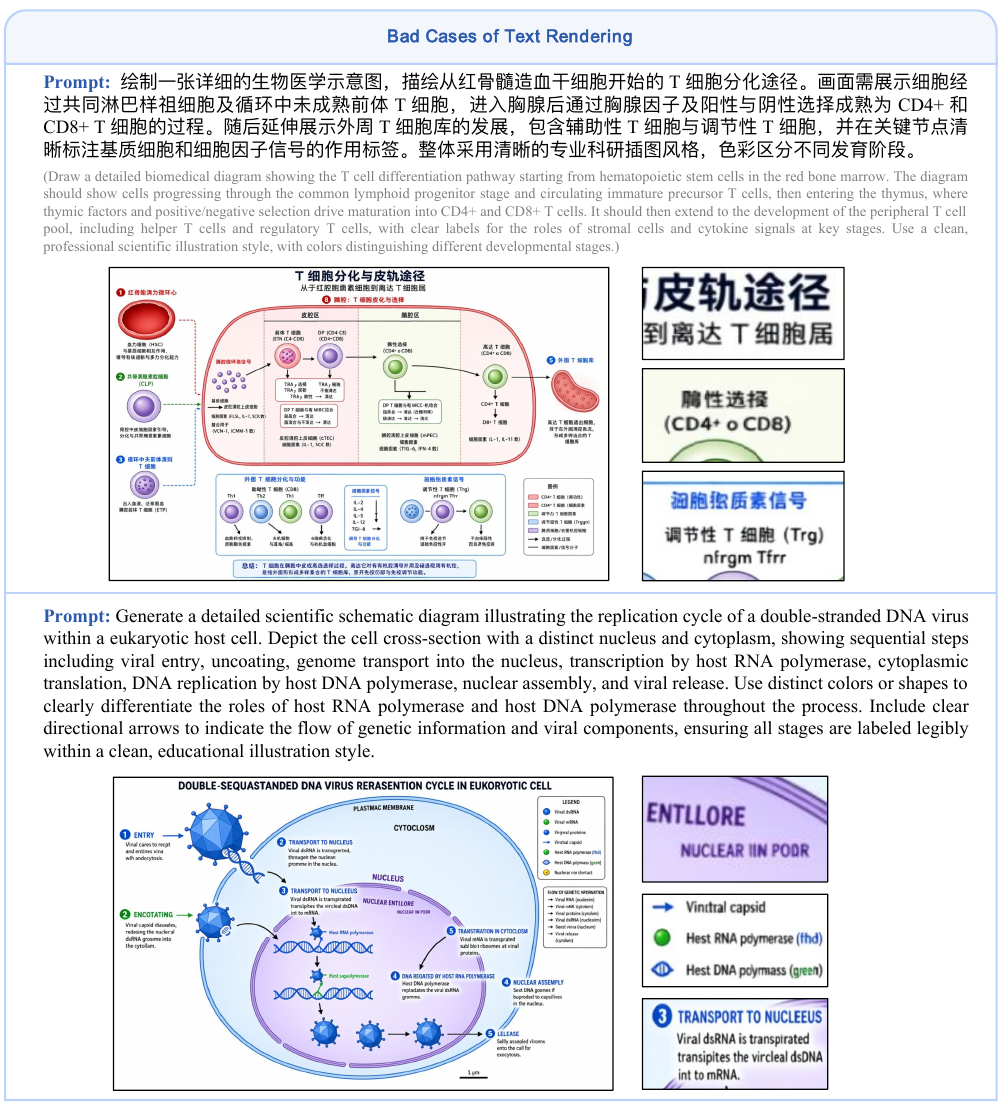}
    \caption{Failure cases in text rendering. The examples show that \model may produce unreadable or incorrect text in scientific images, including glyph-level errors, blurred characters, missing or extra strokes, and semantic substitutions. These failures are especially challenging for dense annotations and Chinese text rendering.}
    \label{fig:limitation_text_rendering}
\end{figure}

\section{Limitations}
\label{sec:limitations}

Although \model achieves strong performance on scientific image generation and editing, several limitations remain. We summarize the main failure modes in text rendering, instruction following, training strategy, and safety.

\paragraph{Text rendering, especially Chinese.}
\model still struggles with long-text and densely annotated scientific images. As shown in Figure~\ref{fig:limitation_text_rendering}, the model may produce glyph-level misspellings, extra or missing strokes, blurred characters, or semantic substitutions. These issues appear in both English and Chinese text rendering, but are more pronounced in Chinese because Chinese characters contain more complex internal structures and because Chinese samples account for a smaller portion of the training data. Such errors can reduce the readability and reliability of generated scientific figures, especially when precise labels, titles, or terminology are required.

\begin{figure}[t!]
    \centering
    \includegraphics[width=0.96\linewidth]{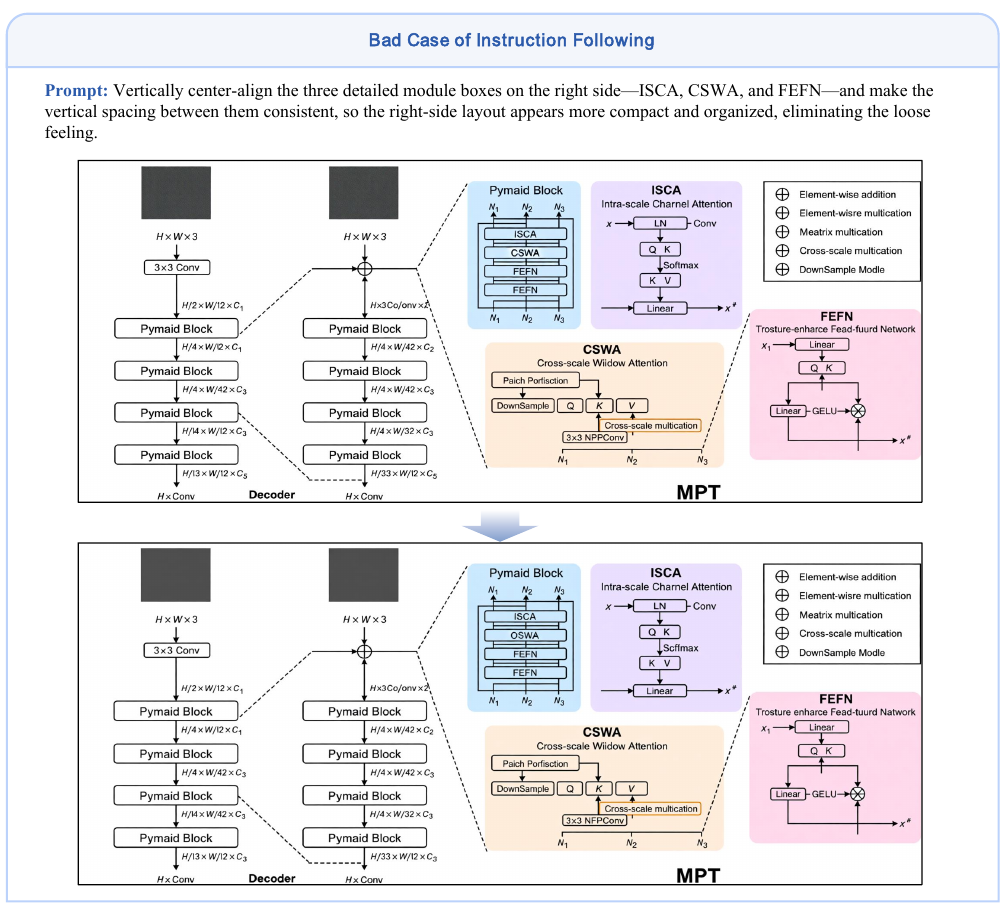}
    \caption{Failure case in instruction following. Given a complex editing instruction, \model fails to apply the requested visual modification and produces an output that remains nearly unchanged from the input image. This suggests that complex local editing and long-horizon instruction execution remain challenging.}
    \label{fig:limitation_instruction_following}
\end{figure}

\paragraph{Instruction following in complex editing.}
The model can perform instruction-based scientific image editing and multi-turn refinement, but it may still fail to execute complex or fine-grained editing instructions. Figure~\ref{fig:limitation_instruction_following} shows a representative failure case where the output image remains nearly unchanged after an editing instruction. This indicates that the model sometimes understands the high-level task but does not sufficiently translate the instruction into effective visual modifications. The problem can become more evident in long-horizon multi-turn editing, where the model must preserve previous edits while applying new local changes.

\paragraph{Training-stage limitations.}
The current model is mainly optimized through supervised fine-tuning and diffusion supervised training. Due to computational and data constraints, we have not conducted large-scale scientific image pre-training, preference optimization, or reinforcement learning for generation and editing. As a result, \model may still lag behind frontier closed-source systems in dense text rendering, fine-grained prompt following, local editing controllability, and aesthetic refinement. Future work will explore stronger pre-training and post-training strategies for reasoning-grounded image generation and editing.

\paragraph{Safety discussion.}
Scientific image generation and editing also introduce safety and compliance risks. Potential misuse includes fabricating experimental figures, producing misleading medical images, modifying scientific evidence, or processing sensitive biomedical data without authorization. Outputs from \model should therefore be treated as assistive generated content rather than clinical diagnosis, treatment evidence, or formal scientific proof. We will provide usage-boundary documentation with the model release and explicitly discourage deceptive scientific presentation, automatic clinical decision-making, and unauthorized use of private or sensitive data.

\section{Conclusion}
\label{sec:conclusion}

We present \model, an open-weight unified multimodal model for scientific image understanding, generation, and editing. Built upon the scientific multimodal reasoning backbone \sonevl, \model adopts a think-before-generate framework that aligns and injects task-oriented reasoning representations into an MMDiT image generation and editing module. This design connects scientific image understanding, generation, and editing within a unified model, making it suitable for detail-sensitive scientific image tasks such as scientific illustration generation, scientific image segmentation, medical image translation, and super-resolution.

We construct SciGenEdit, a 314K-sample training dataset covering scientific image generation, scientific image editing, and scientific image understanding, and release the model weights, inference service code, and SciGenEdit-10K under the Apache 2.0 license. By formulating segmentation, translation, and super-resolution as native image editing tasks, \model handles multiple classes of scientific image editing tasks within a unified framework. Experiments show that \model outperforms open-source baselines in scientific illustration generation, achieves state-of-the-art results on four scientific image editing benchmarks, and largely preserves the scientific multimodal understanding capability inherited from \sonevl.

Despite these results, scientific image generation and editing remain challenging. \model still shows limitations in dense text rendering, fine-grained local editing, and complex instruction following. Future work will expand scientific image data, strengthen multi-turn and local editing, improve text rendering and domain verification, and extend the framework to additional scientific modalities such as protein structures, spectral data, and material chemical formulas.

\section*{Author Contributions}

All contributors of this paper are listed below.

\textbf{Contributors:}
Qingxiao Li,
Zikai Wang,
Qingli Wang,
Nan Xu.

\bibliographystyle{colm2024_conference}
\bibliography{references}

\end{document}